\documentclass[11pt]{article}

\usepackage{booktabs}
\usepackage{colortbl}
\usepackage[table]{xcolor}
\definecolor{lightblue}{RGB}{220,235,250}
\definecolor{boxframe}{RGB}{175,210,240} 
\definecolor{boxback}{RGB}{242, 242, 245} 
\definecolor{systemcolor}{RGB}{0, 90, 160}    
\definecolor{usercolor}{RGB}{0, 140, 70}      
\definecolor{assistantcolor}{RGB}{130, 50, 200} 
\usepackage{amsmath}
\usepackage{amssymb}
\usepackage{xcolor}
\usepackage{pifont}

\usepackage{subcaption}
\usepackage{algorithm}
\usepackage{algorithmic}
\usepackage{dsfont}
\usepackage{enumitem}
\usepackage[most]{tcolorbox} 
\usepackage{tabularx}
\usepackage{multirow}

\usepackage{marvosym}
\usepackage{footmisc}
\usepackage[final]{acl}

\usepackage{times}
\usepackage{latexsym}

\usepackage[T1]{fontenc}

\usepackage[utf8]{inputenc}

\usepackage{microtype}

\usepackage{inconsolata}

\usepackage{graphicx}

\usepackage{multirow}

\title{DySem: Uncovering Dynamic Semantic Components of Large Language Models for Calculating Semantic Textual Similarity}

\author{
    Kaijie Zheng$^{*}$,
    Weiqin Wang$^{*}$,
    Yile Wang\textsuperscript{\Letter},
    Hui Huang \\
    Guangdong Provincial Key Laboratory of Visual Media and Multidimensional Intelligence\\CSSE, Shenzhen University \\
}

\begin{document}
\maketitle

\DefineFNsymbols*{authsymbols}{*\Letter}
\setfnsymbol{authsymbols}
\renewcommand{\thefootnote}{\fnsymbol{footnote}}
\footnotetext[1]{Equal contribution.}
\footnotetext[2]{Corresponding to \textit{wangyile@szu.edu.cn}.}
\renewcommand{\thefootnote}{\arabic{footnote}}
\setcounter{footnote}{0}

\begin{abstract}
Calculating semantic textual similarity is a foundational task in natural language processing. Current large language models (LLMs) based methods typically rely on extracting last-layer hidden states with fixed dimensions to compute similarity for every text pairs. We argue that this paradigm is suffer from two limitations: (i) The last hidden layer encodes more general knowledge rather than just semantic knowledge, making it suboptimal for semantic similarity computation; (ii) The hidden layer dimensions of LLMs are generally very large, which introduces some redundancy and noise for representing semantics. In this work, we propose \textsc{DySem}, a novel training-free framework that investigates more semantic-related internal components of LLMs via \textit{multilingual consensus}, and shifts away from static representation spaces in favor of dynamic, sample-specific semantic dimensions by constructing text-dependent joint semantic set and computes similarity over this shared dimensional subset. Extensive experiments across various LLMs show that our method consistently outperforms recent baselines while maintaining lower dimensions for similarity calculation. The code is released at \href{https://github.com/szu-tera/DySem}{\texttt{https://github.com/szu-tera/DySem}}.

\end{abstract}




\section{Introduction}

\begin{figure}[t!]
  \centering
  \includegraphics[width=\columnwidth]{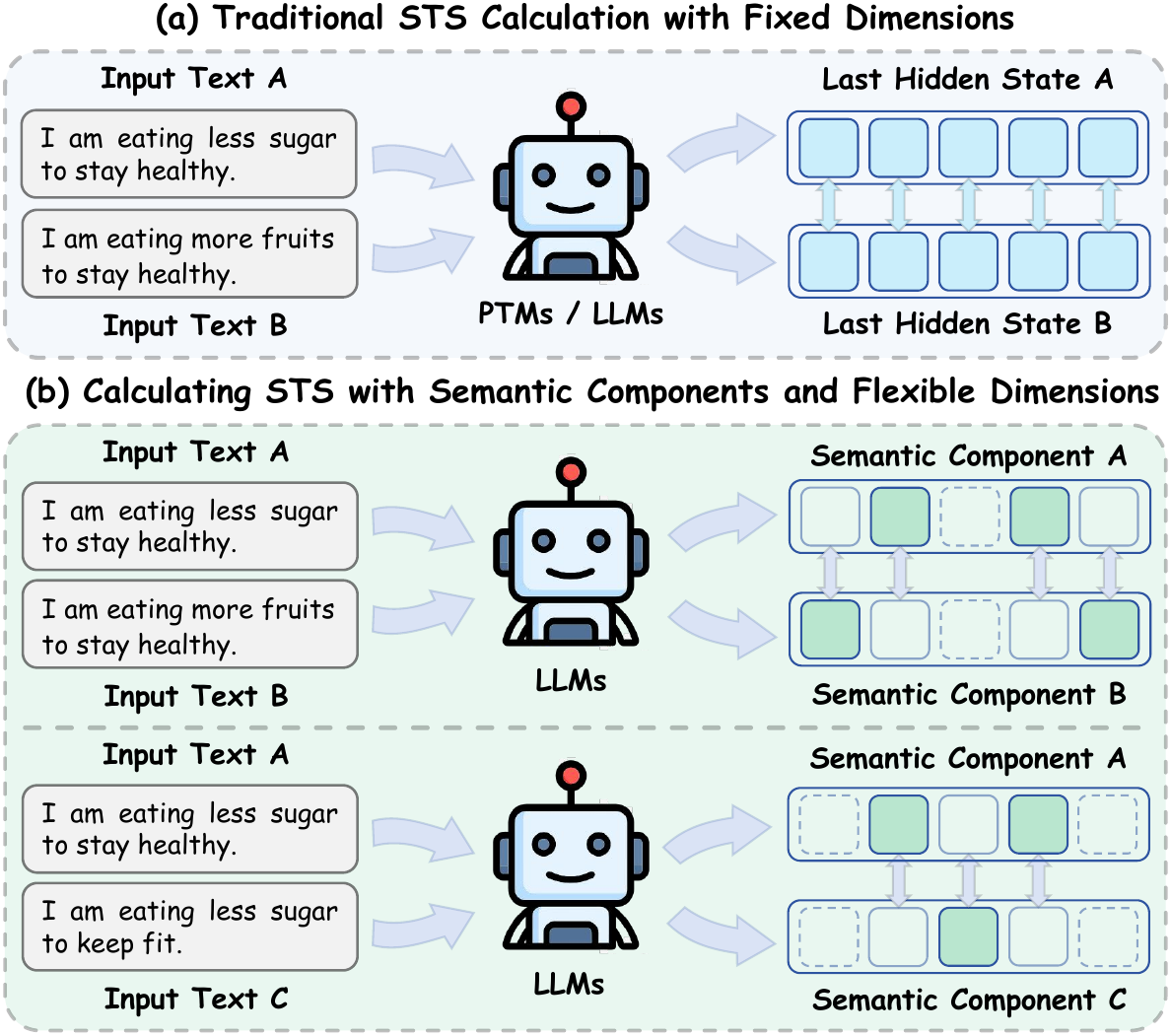}
  \caption{Illustration of different ways for calculating semantic similarity: (a) conventional methods use last hidden states and full dimensions; (b) our \textsc{DySem} investigates semantic components and flexible dimensions.}
  \label{fig:intro}
\end{figure}

Evaluating the relationship between two texts and calculating their semantic textual similarity (STS;~\citealp{agirre-etal-2012-semeval}) plays an important role in natural language processing, serving as the backbones of applications such as semantic search, clustering, and information retrieval. 
Conventional pre-trained models~(PTMs; \citealp{devlin-etal-2019-bert, reimers-gurevych-2019-sentence, gao-etal-2021-simcse}) and large language models~(LLMs; \citealp{prompteol,metaeol,cheng-etal-2025-contrastive}) based methods typically map \textit{all} text into a vector using the \textit{same} last-layer hidden states with \textit{fixed} dimensions for similarity calculation, as shown in Figure~\ref{fig:intro}(a).

However, we argue that this default design has two limitations: First, the last-layer hidden states of LLMs typically encodes highly generalized information optimized for next-token prediction, which inevitably entangles core semantics with broader unrelated information. Studies also show that semantics are also captured by internal neurons of LLMs~\cite{skean2025layer, zhang2026llm}. This motivates us to explore direct internal semantic components of LLMs for similarity computation.

Second, representing texts using the entire high-dimensional states can introduces information redundancy, and the semantic integrity is largely preserved after dimensionality reduction~\cite{wang-etal-2023-dimensionality, inkiriwang-etal-2025-really, wang-etal-2025-ldir}.  Considering that the hidden state sizes of LLMs are often very large, this brings unnecessary overhead in textual similarity computation and downstream applications. This motivates us to consider using a more flexible number of dimensions to compute the STS between two texts.

In this work, we propose \textsc{DySem}, a training-free method that aims to extract semantic components
of LLMs beyond last-layer hidden states and shifts away from fixed representation dimensions in favor of dynamic, sample-specific semantic dimensions, as shown in Figure~\ref{fig:intro}(b). In particular, \textsc{DySem} first extracts a sample-specific set of semantic components and dimensions via \emph{multilingual consensus}, drawing from the straightforward intuition that 
a text's core semantic meaning remains invariant across its translated versions. Then, \textsc{DySem} dynamically construct a pair-dependent \emph{joint semantic set} by taking the union of the semantic dimensions from both texts, and compute the STS over this joint semantic set. 
This dynamic alignment effectively filters out irrelevant background noise, ensuring that the metric evaluates the shared semantic concepts between texts.

Experiments on seven standard STS benchmarks across various LLMs show that, \textsc{DySem} can effectively uncovers attention-based semantic components of LLMs and demonstrates superior STS performance compared with recent baselines without model training. Moreover, \textsc{DySem} maintains flexible computational dimensions and can achieve competitive results using only around 1,000 dimensions or fewer on average across all samples.

\section{Preliminaries and Notations}


\subsection{Internal Components of Transformer}
\label{subsec:decoder_components}

Consider a decoder-only Transformer model $\mathcal{M}$ with layers $L$ and hidden size $d$. Let $N$ denotes the sequence length and $h_n^l \in \mathbb{R}^{d}$ denotes the hidden state of token $n$ at layer $l$, where $n\in\{0,1,\dots,N\}$, $l\in\{0,1,\dots,L\}$.

At layer $l$, the input state $h_n^{l-1}$ is first processed by the multi-head attention module and the attention output $a_n^l$ is calculated by:
\begin{equation}
    a_n^l = \mathrm{MHA}^l(h_n^{l-1}),
\end{equation}
where $\mathrm{MHA}^l$ denotes masked multi-head self-attention computed by the standard query-key-value mechanism. The resulting attention output $a_n^l$ is added by residual connection, and gives the output of feedforward network $f_n^l$:
\begin{equation}
    f_n^l = \mathrm{FFN}^l(h_n^{l-1} + a_n^l).
\end{equation}

The final hidden state $h_n^l$ at layer $l$ is calculated by the summation of the attention output, FFN output, and the hidden states from previous layer:
\begin{equation}
    h_n^l = h_n^{l-1} + a_n^l + f_n^l,
\end{equation}
where $a_n^l$ and $f_n^l$ represent the contributions of the MHA and FFN at layer $l$, respectively. Expanding the residual updates over all layers yields:
\begin{equation}
    h_n^L = h_n^0 + \sum_{l=1}^{L} a_n^l + \sum_{l=1}^{L} f_n^l.
    \label{eq:att_ffn}
\end{equation}

For convenience, define the cumulative attention and FFN contributions as:
\begin{equation}
    A_n^L = \sum_{l=1}^{L} a_n^l,
    \qquad
    F_n^L = \sum_{l=1}^{L} f_n^l.
    \label{eq:cum_att_ffn}
\end{equation}

Thereby, the final-layer hidden state can be re-written equivalently as:
\begin{equation}
    h_n^L = h_n^0 + A_n^L + F_n^L.
    \label{eq:hiddenstates}
\end{equation}

For simplicity and clarity in our mechanistic analysis, we omit Layer Normalization (LayerNorm or RMSNorm) terms from these equations. 

\begin{figure*}[t!]
  \centering
  \includegraphics[width=1\textwidth]{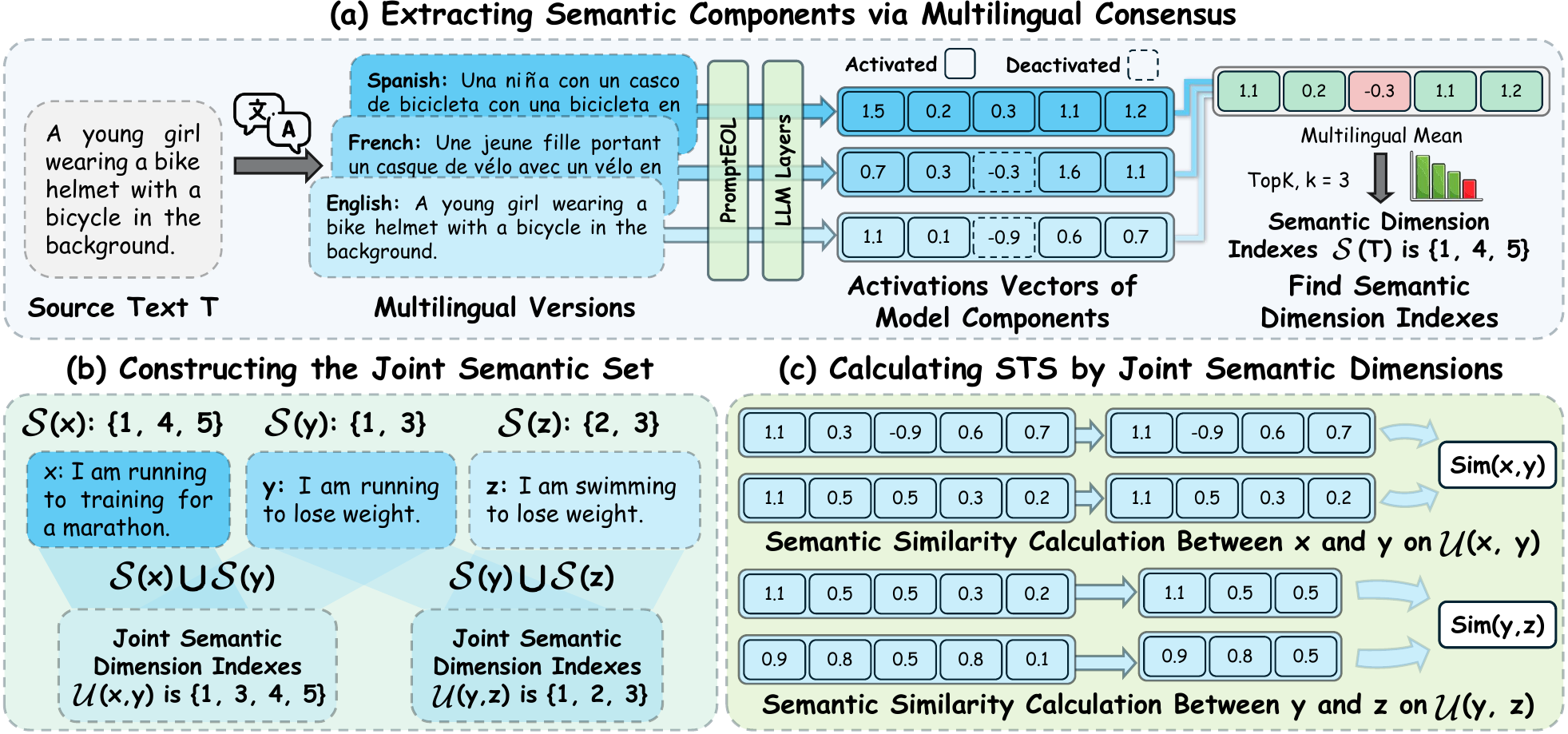}
  \caption{Overview of \textsc{DySem}. The process involves (a) \textbf{extracting} semantic component dimension via multilingual consensus (\S\ref{subsec:method_dim_extract}), (b) \textbf{constructing} a pair-dependent joint semantic set $\mathcal{U}(x,y)$ by merging the individual semantic component dimension sets (\S\ref{subsec:method_set_constru}), and (c) \textbf{calculating} the semantic similarity in the selected joint dimensions (\S\ref{subsec:method_sim_calc}).}
  \label{fig:method}
\end{figure*}

\subsection{Semantic Textual Similarity Calculation}

Given text pair $(x, y)$, conventional encoder-based methods such as SimCSE~\cite{gao-etal-2021-simcse} perform contrastive learning with the hidden state $h_n^L$ of special token (e.g., [CLS]) as text representation and calculate their similarity using cosine distance:
\begin{equation}
    s(x,y)=\cos(h_n^L(x),h_n^L(y)),
\end{equation}
where $h_n^L$ can also be replaced by the pooling results of all hidden states $h_0^L,\ldots,h_n^L$.

To extract the text meaning from decoder-based LLMs without training, PromptEOL~\cite{prompteol} wraps each text $x$ with template $\mathcal{T}$ and extract the last hidden state $h_N^L$ as text representation:
\begin{equation}
    \mathcal{M}(\mathcal{T}(x))=[h_0^L,\ldots,h_N^L],
\end{equation}
where $\mathcal{T}$ is used to extract text meanings as follows:
\begin{equation}
\scalebox{0.9}{$
\mathcal{T} = \text{\textit{This sentence: ``$x$'' means in one word:}}
$}
\label{eq:template}
\end{equation}

\paragraph{Our Motivations.}
Whether using the hidden states from the special token or the last token, the above methods both use a fixed hidden state to represent the semantics of any given text $x$, and they utilize the full dimensionality of the states, i.e., the pre-defined hidden size $d$, when computing the semantic similarity. This motivates us to explore the following two questions. (1) \textbf{Representation Selection}: Can alternative internal components such as attention or FFN vectors in Eq.~\ref{eq:att_ffn} and Eq.~\ref{eq:cum_att_ffn} be used for better representing semantics? (2) \textbf{Dimension Selection}: Do we really need two $d$-dimensional (typically 4096 or more for LLMs) vectors to compute textual similarity?

\section{Method}

Our method \textsc{DySem} proceeds in three stages: (i) extracting \emph{semantic components and dimensions} for each text via \emph{multilingual consensus} (\S\ref{subsec:method_dim_extract}), (ii) constructing a pair-dependent \emph{joint semantic set} by merging these individual dimensions (\S\ref{subsec:method_set_constru}), and (iii) computing a \emph{restricted cosine similarity} over the union of these dimension sets (\S\ref{subsec:method_sim_calc}). The overall framework is shown in Figure~\ref{fig:method}.

\subsection{Extracting Semantic Components}
\label{subsec:method_dim_extract}

Given a text $x$, we first translate it into a predefined set of $m$ target languages $\mathcal{L} = \{\ell_1, \dots, \ell_m\}$, yielding a collection of language-specific renderings:
\begin{equation}
    x^{(\ell)} = T_\ell(x), \quad \ell \in \mathcal{L},
    \label{eq:languageset}
\end{equation}
where $T_\ell$ is the translator with target language $\ell$.

To extract the semantic representation of $x$, we wrap each language version $x^{(\ell)}$ in the prompt template $\mathcal{T}$ to get their corresponding representations. We consider two prompt strategies: (1) \textbf{English Prompt}, which uses the same English template in Eq.~\ref{eq:template} as the prompt for LLMs, and (2) \textbf{Language-Specific Prompt}, which uses translated version of the template $\mathcal{T}$ using the translator $T_\ell$ as the prompt. Then, we can get the internal state vector $\mathbf{s}^{(\ell)}(x) \in \mathbb{R}^d$ from the last-token, constructing a set of multilingual text vectors $\mathcal{E}(x) = \{\mathbf{s}^{(\ell)}(x) \mid \ell \in \mathcal{L}\}$. For the internal state vector $\mathbf{s}^{(\ell)}(x) \in \mathbb{R}^d$, we consider three components  $h_n^L$, $A_n^L$, and $F_n^L$ as defined in \S\ref{subsec:decoder_components}. We analyze these components in \S\ref{subsec:analysis_component_selection}.

\newcommand{\modelname}[1]{\rotatebox[origin=c]{90}{\scriptsize{#1}}}
\newcommand{\wmethod}[1]{{\textit{w}/ #1}}
\newcommand{\ourscell}[1]{\cellcolor{lightblue!100}#1}

\begin{table*}[t!]
  \centering
  \small
  \resizebox{\textwidth}{!}{%
    \begin{tabular}{c l c *{8}{c}}
      \toprule
      \multicolumn{2}{c}{\textbf{Methods}} & \textbf{\#Dim.} & \textbf{STS12} & \textbf{STS13} & \textbf{STS14} & \textbf{STS15} & \textbf{STS16} & \textbf{\mbox{STS-B}} & \textbf{\mbox{SICK-R}} & \textbf{Avg.} \\
      \midrule
      \multicolumn{2}{l}{{Sentence-BERT (110M)}$^\dagger$~\cite{reimers-gurevych-2019-sentence}} & $\phantom{^{*}}768\phantom{^{*}}$ & 70.97 & 76.53 & 73.19 & 79.09 & 74.30 & 77.03 & 72.91 & 74.89 \\
      \multicolumn{2}{l}{{SimCSE (110M)}$^\dagger$~\cite{gao-etal-2021-simcse}} & $\phantom{^{*}}768\phantom{^{*}}$ & 68.40 & 82.41 & 74.38 & 80.91 & 78.56 & 76.85 & 72.23 & 76.25 \\
      \multicolumn{2}{l}{{PromptBERT (110M)}$^\dagger$~\cite{jiang-etal-2022-promptbert}} & $\phantom{^{*}}768\phantom{^{*}}$ & 71.56 & 84.58 & 76.98 & 84.47 & 80.60 & 81.60 & 69.87 & 78.54 \\
      \midrule
      \multirow{10}{*}{\modelname{LLaMA2-7B}} 
      & {Mean Pooling} & $\phantom{^{*}}4096\phantom{^{*}}$ & 35.43 & 53.63 & 39.88 & 54.69 & 52.99 & 42.66 & 49.45 & 46.96 \\
      & {PromptEOL}$^\dagger$~\cite{prompteol} & $\phantom{^{*}}4096\phantom{^{*}}$ & 58.80 & 77.00 & 66.34 & 73.22 & 73.53 & 71.72 & 70.19 & 70.11 \\
      & {LLM2Vec}~\cite{behnamghader2024llmvec} & $\phantom{^{*}}4096\phantom{^{*}}$ & 65.39 & 79.26 & 72.98 & 82.72 & 81.02 & 78.32 & 71.77 & 75.92 \\
      & {Token Prepending}$^\dagger$~\cite{fu-etal-2025-token} & $\phantom{^{*}}4096\phantom{^{*}}$ & 66.90 & 83.12 & 74.31 & 79.87 & \textbf{80.03} & \textbf{80.67} & \underline{75.40} & 77.19 \\
      & {Contrastive Prompting}$^\dagger$(Scaling;~\citealp{cheng-etal-2025-contrastive}) & $\phantom{^{*}}4096\phantom{^{*}}$ & 63.34 & 82.15 & 71.73 & 79.68 & 77.23 & 78.71 & 74.04 & 75.27 \\
      & {Contrastive Prompting}$^\dagger$(Recovering;~\citealp{cheng-etal-2025-contrastive}) & $\phantom{^{*}}4096\phantom{^{*}}$ & 63.37 & 81.95 & 71.90 & 79.54 & 77.29 & 78.36 & 74.00 & 75.20 \\
      & {SemPA}~\cite{chen2026sempa} & $\phantom{^{*}}4096\phantom{^{*}}$ & 68.77 & 83.73 & 74.73 & 82.08 & 77.48 & 79.76 & \textbf{77.26} & 77.69 \\
      & {AlignedWVA}~\cite{zhang2026llm} & $\phantom{^{*}}4096\phantom{^{*}}$ & 67.94 & \underline{84.97} & 75.93 & 81.68 & \underline{79.98} & \underline{80.39} & 74.45 & 77.91 \\
      & \ourscell{\textbf{\textsc{DySem} (English Prompt, Ours)}} & \ourscell{$\phantom{^{*}}1238^{*}$} & \ourscell{\underline{70.20}} & \ourscell{\textbf{85.09}} & \ourscell{\textbf{77.39}} & \ourscell{\underline{82.64}} & \ourscell{79.73} & \ourscell{\textbf{80.67}} & \ourscell{74.75} & \ourscell{\underline{78.64}} \\  
      & \ourscell{\textbf{\textsc{DySem} (Language-specific Prompt, Ours)}} & \ourscell{$\phantom{^{*}}1240^{*}$} & \ourscell{\textbf{72.68}} & \ourscell{84.75} & \ourscell{\underline{77.28}} & \ourscell{\textbf{83.47}} & \ourscell{79.19} & \ourscell{80.06} & \ourscell{75.39} & \ourscell{\textbf{78.98}} \\
      \midrule
      \multirow{5}{*}{\modelname{LLaMA3.1-8B}} 
      & {Mean Pooling} & $\phantom{^{*}}4096\phantom{^{*}}$ & 35.33 & 53.63 & 43.63 & 56.74 & 53.27 & 45.90 & 50.75 & 48.46 \\
      & {PromptEOL}~\cite{prompteol} & $\phantom{^{*}}4096\phantom{^{*}}$ & 60.77 & 80.16 & 69.64 & 76.90 & 78.23 & 72.74 & 70.45 & 72.70 \\
      & {AlignedWVA}~\cite{zhang2026llm} & $\phantom{^{*}}4096\phantom{^{*}}$ & 64.13 & \underline{83.98} & 72.66 & 80.41 & 80.64 & 78.91 & 72.99 & 76.25 \\
      & \ourscell{\textbf{\textsc{DySem} (English Prompt, Ours)}} & \ourscell{$\phantom{^{*}}1103^{*}$} & \ourscell{\underline{66.75}} & \ourscell{83.70} & \ourscell{\underline{73.41}} & \ourscell{\underline{81.47}} & \ourscell{\underline{80.70}} & \ourscell{\underline{79.49}} & \ourscell{\underline{74.23}} & \ourscell{\underline{77.11}} \\
      & \ourscell{\textbf{\textsc{DySem} (Language-specific Prompt, Ours)}} & \ourscell{$\phantom{^{*}}995^{*}$} & \ourscell{\textbf{74.95}} & \ourscell{\textbf{84.18}} & \ourscell{\textbf{76.22}} & \ourscell{\textbf{83.02}} & \ourscell{\textbf{81.30}} & \ourscell{\textbf{81.79}} & \ourscell{\textbf{77.25}} & \ourscell{\textbf{79.82}} \\
      \midrule
      \multirow{5}{*}{\modelname{Qwen2.5-7B}} 
      & {Mean Pooling} & $\phantom{^{*}}3584\phantom{^{*}}$ & 37.62 & 44.31 & 32.01 & 43.36 & 46.62 & 33.84 & 42.49 & 40.04 \\
      & {PromptEOL}~\cite{prompteol} & $\phantom{^{*}}3584\phantom{^{*}}$ & 65.50 & 79.70 & 70.62 & 79.05 & 78.73 & 76.44 & 70.46 & 74.36 \\
      & {AlignedWVA}~\cite{zhang2026llm} & $\phantom{^{*}}3584\phantom{^{*}}$ & 71.56 & 80.19 & 69.61 & 75.63 & 78.69 & 74.45 & 73.06 & 74.74 \\
      & \ourscell{\textbf{\textsc{DySem} (English Prompt, Ours)}} & \ourscell{$\phantom{^{*}}978^{*}$} & \ourscell{\textbf{76.92}} & \ourscell{\textbf{83.20}} & \ourscell{\textbf{75.48}} & \ourscell{\textbf{83.51}} & \ourscell{\underline{79.57}} & \ourscell{\textbf{80.29}} & \ourscell{\textbf{78.43}} & \ourscell{\textbf{79.63}} \\
      & \ourscell{\textbf{\textsc{DySem} (Language-Specific Prompt, Ours)}} & \ourscell{$\phantom{^{*}}940^{*}$} & \ourscell{\underline{75.12}} & \ourscell{\underline{82.66}} & \ourscell{\underline{73.94}} & \ourscell{\underline{81.74}} & \ourscell{\textbf{79.80}} & \ourscell{\underline{78.42}} & \ourscell{\underline{77.94}} & \ourscell{\underline{78.52}} \\
      \midrule
      \multirow{5}{*}{\modelname{Qwen3-8B}} 
      & {Mean Pooling} & $\phantom{^{*}}4096\phantom{^{*}}$ & 44.04 & 38.44 & 31.63 & 46.34 & 43.39 & 29.86 & 38.04 & 38.82 \\
      & {PromptEOL}~\cite{prompteol} & $\phantom{^{*}}4096\phantom{^{*}}$ & 59.30 & 74.19 & 61.56 & 72.36 & 74.64 & 73.41 & 66.90 & 68.91 \\
      & {AlignedWVA}~\cite{zhang2026llm} & $\phantom{^{*}}4096\phantom{^{*}}$ & 64.91 & 80.90 & 70.56 & 77.38 & 78.90 & 76.15 & 74.75 & 74.79 \\
      & \ourscell{\textbf{\textsc{DySem} (English Prompt, Ours)}} & \ourscell{$\phantom{^{*}}1027^{*}$} & \ourscell{\underline{75.22}} & \ourscell{\underline{83.22}} & \ourscell{\underline{76.82}} & \ourscell{\underline{83.49}} & \ourscell{\underline{79.54}} & \ourscell{\underline{80.09}} & \ourscell{\textbf{79.16}} & \ourscell{\underline{79.65}} \\
      & \ourscell{\textbf{\textsc{DySem} (Language-Specific Prompt, Ours)}} & \ourscell{$\phantom{^{*}}675^{*}$} & \ourscell{\textbf{77.54}} & \ourscell{\textbf{84.33}} & \ourscell{\textbf{76.95}} & \ourscell{\textbf{83.91}} & \ourscell{\textbf{81.19}} & \ourscell{\textbf{81.26}} & \ourscell{\underline{78.56}} & \ourscell{\textbf{80.54}} \\
      \bottomrule
    \end{tabular}%
  }
  \caption{Comparison between baselines and our method on base models. In each column, the best results are in \textbf{bold}, and the second-best results are \underline{underlined}. Our method's performance is highlighted in blue. $^\dagger$ denotes results from the original paper. $^*$ indicates the average dimension of vectors across all samples for similarity computation.}
  \label{tab:main_exp_base}
\end{table*}

We define \emph{multilingual consensus} as the consistent activation patterns shared across these diverse linguistic renderings. By leveraging this consensus, we can effectively filter out language-specific surface artifacts and isolate the core text-relevant semantics. Specifically, this consensus is operationalized by identifying dimensions that are consistently activated across all language versions. Assuming a dimension is activated when its value is positive~\cite{tang-etal-2024-language, neo-etal-2024-interpreting, lee-etal-2025-small}, the consensus index set is defined as:
\begin{equation}
    \mathcal{I}(x) = \left\{ j \;\middle|\; s_j^{(\ell)}(x) > 0, \; \forall\, \ell \in \mathcal{L} \right\},
\end{equation}
where $s_j^{(\ell)}(x)$ is the $j$-th component of $\mathbf{s}^{(\ell)}(x)$.

Among the co-activated dimensions $\mathcal{I}(x)$, we rank each dimension by its mean activation across all languages:
\begin{equation}
    \bar{s}_j(x) = \frac{1}{m} \sum_{\ell \in \mathcal{L}} s_j^{(\ell)}(x), \quad j \in \mathcal{I}(x),
\end{equation}
and retain the top-$k$ dimensions with the highest mean activation:
\begin{equation}
    \mathcal{S}(x) =
    \operatorname{TopKIdx}\!\left(\left\{\bar{s}_j(x)\right\}_{j \in \mathcal{I}(x)}\right),
    \label{eq:topk}
\end{equation}
where $\operatorname{TopKIdx}$ returns indices of the top $k$ values.

The index set $\mathcal{S}(x) \subset \{1, \dots, d\}$ constitutes the \emph{semantic component set} of text $x$: a compact summary of the dimensions most reliably encoding its meaning based on the multilingual consensus. 
An analysis of the choice of $k$ is provided in \S\ref{subsec:analysis_dimension_size}.

\subsection{Constructing the Joint Semantic Set}
\label{subsec:method_set_constru}
The size of index set $\mathcal{S}(x)$ depends on the text $x$. For a given text pair $(x, y)$, to facilitate calculating their similarity practically, we construct \emph{joint semantic set} as the union of their index set $\mathcal{S}(x)$ and $\mathcal{S}(y)$ from their semantic components:
\begin{equation}
    \mathcal{U}(x, y) = \mathcal{S}(x) \cup \mathcal{S}(y).
\end{equation}

This union preserves dimensions important to either text, allowing the comparison to capture both semantic overlap and difference. Since $\mathcal{U}(x,y)$ is derived directly from the text-specific sets $\mathcal{S}(x)$ and $\mathcal{S}(y)$, it is inherently determined dynamically for each text pair.

\subsection{Calculating by Joint Semantic Dimensions}
\label{subsec:method_sim_calc}

For any text $z$, let $\mathbf{v}(z)$ denote the internal state vector from the model. We further consider two variants of such vector: (i) using the source-language vector, and (ii) using the multilingual mean vector:
\begin{equation}
\mathbf{v}(z)=
\begin{cases}
\mathbf{s}^{(\ell_z)}(z), & \text{source},\\
\tfrac{1}{m}\sum_{\ell\in\mathcal{L}}\mathbf{s}^{(\ell)}(z), & \text{mean},
\end{cases}
\label{eq:vector}
\end{equation}
where $\ell_z$ is the source language of $z$.

We then construct text-specific joint semantic representation on $\mathcal{U}(x,y)$:
\begin{equation}
    \mathbf{v}_{\mathcal{U}}(z)=\mathbf{v}(z)\big|_{\mathcal{U}(x,y)}, \quad z\in\{x,y\},
\end{equation}
where $\mathbf{v}(z)\big|_{\mathcal{U}(x,y)}$ selects the components indexed by $\mathcal{U}(x,y)$. The cosine similarity is computed as:
\begin{equation}
    \operatorname{sim}(x, y)=
    \cos\!\left(\mathbf{v}_{\mathcal{U}}(x), \mathbf{v}_{\mathcal{U}}(y)\right).
\end{equation}

Considering that the dimensions of vectors $\mathbf{v}_{\mathcal{U}}(x)$ and $\mathbf{v}_{\mathcal{U}}(y)$ are determined by the joint semantic set $\mathcal{U}(x,y)$ and depend on the input texts $x$ and $y$, thus the cosine similarity is computed over flexible dimensions.

\begin{table*}[t!]
  \centering
  \small
  \resizebox{\textwidth}{!}{%
    \begin{tabular}{c l c *{8}{c}}
      \toprule
      \multicolumn{2}{c}{\textbf{Methods}} & \textbf{\#Dim.} & \textbf{STS12} & \textbf{STS13} & \textbf{STS14} & \textbf{STS15} & \textbf{STS16} & \textbf{\mbox{STS-B}} & \textbf{\mbox{SICK-R}} & \textbf{Avg.} \\
      \midrule
      
      \multirow{5}{*}{\modelname{LLaMA2-7B-it}} 
      & {Mean Pooling} & $\phantom{^{*}}4096\phantom{^{*}}$ & 41.07 & 58.26 & 47.07 & 61.63 & 58.99 & 48.42 & 53.16 & 52.66 \\
      & {PromptEOL}~\cite{prompteol} & $\phantom{^{*}}4096\phantom{^{*}}$ & 59.19 & 78.18 & 67.77 & 75.57 & 71.35 & 72.69 & 70.45 & 70.74 \\
      & {AlignedWVA}~\cite{zhang2026llm} & $\phantom{^{*}}4096\phantom{^{*}}$ & 60.95 & \underline{79.83} & 69.90 & 80.56 & \underline{75.57} & 76.97 & 75.58 & 74.20 \\
      & \ourscell{\textbf{\textsc{DySem} (English Prompt, Ours)}} & \ourscell{$\phantom{^{*}}968^{*}$} & \ourscell{\underline{64.07}} & \ourscell{79.62} & \ourscell{\underline{71.91}} & \ourscell{\underline{82.04}} & \ourscell{\textbf{76.01}} & \ourscell{\underline{77.99}} & \ourscell{\underline{75.99}} & \ourscell{\underline{75.38}} \\
      & \ourscell{\textbf{\textsc{DySem} (Language-Specific Prompt, Ours)}} & \ourscell{$\phantom{^{*}}1193^{*}$} & \ourscell{\textbf{70.84}} & \ourscell{\textbf{80.52}} & \ourscell{\textbf{74.85}} & \ourscell{\textbf{82.69}} & \ourscell{74.61} & \ourscell{\textbf{79.23}} & \ourscell{\textbf{77.39}} & \ourscell{\textbf{77.16}} \\
      \midrule
      \multirow{5}{*}{\modelname{LLaMA3.1-8B-it}} 
      & {Mean Pooling} & $\phantom{^{*}}4096\phantom{^{*}}$ & 38.65 & 55.28 & 45.32 & 60.04 & 56.36 & 48.23 & 51.96 & 50.84 \\
      & {PromptEOL}~\cite{prompteol} & $\phantom{^{*}}4096\phantom{^{*}}$ & 62.27 & 78.30 & 69.53 & 76.32 & 78.80 & 74.99 & 71.97 & 73.17 \\
      & {AlignedWVA}~\cite{zhang2026llm} & $\phantom{^{*}}4096\phantom{^{*}}$ & 64.89 & 80.89 & 71.97 & 81.04 & \underline{82.35} & 80.28 & 75.67 & 76.73 \\
      & \ourscell{\textbf{\textsc{DySem} (English Prompt, Ours)}} & \ourscell{$\phantom{^{*}}1176^{*}$} & \ourscell{\underline{74.41}} & \ourscell{\underline{83.62}} & \ourscell{\underline{76.11}} & \ourscell{\underline{82.77}} & \ourscell{81.01} & \ourscell{\underline{82.56}} & \ourscell{\underline{77.84}} & \ourscell{\underline{79.76}} \\
      & \ourscell{\textbf{\textsc{DySem} (Language-Specific Prompt, Ours)}} & \ourscell{$\phantom{^{*}}772^{*}$}  & \ourscell{\textbf{76.38}} & \ourscell{\textbf{84.87}} & \ourscell{\textbf{77.23}} & \ourscell{\textbf{83.92}} & \ourscell{\textbf{82.50}} & \ourscell{\textbf{83.77}} & \ourscell{\textbf{79.74}} & \ourscell{\textbf{81.20}} \\
      \midrule
      \multirow{5}{*}{\modelname{Qwen2.5-7B-it}} 
      & {Mean Pooling} & $\phantom{^{*}}3584\phantom{^{*}}$ & 41.56 & 45.39 & 33.67 & 44.40 & 50.63 & 35.35 & 43.60 & 42.09 \\
      & {PromptEOL}~\cite{prompteol} & $\phantom{^{*}}3584\phantom{^{*}}$ & 66.90 & 78.21 & 69.38 & 77.24 & 76.05 & 74.65 & 67.02 & 72.78 \\
      & {AlignedWVA}~\cite{zhang2026llm} & $\phantom{^{*}}3584\phantom{^{*}}$ & 68.20 & 77.49 & 68.47 & 72.73 & 75.26 & 74.35 & 67.40 & 71.98 \\
      & \ourscell{\textbf{\textsc{DySem} (English Prompt, Ours)}} & \ourscell{$\phantom{^{*}}1001^{*}$} & \ourscell{\textbf{77.54}} & \ourscell{\textbf{82.28}} & \ourscell{\textbf{75.35}} & \ourscell{\textbf{83.49}} & \ourscell{\underline{78.89}} & \ourscell{\textbf{80.49}} & \ourscell{\underline{78.08}} & \ourscell{\textbf{79.44}} \\
      & \ourscell{\textbf{\textsc{DySem} (Language-Specific Prompt, Ours)}} & \ourscell{$\phantom{^{*}}900^{*}$} & \ourscell{\underline{75.12}} & \ourscell{\underline{82.03}} & \ourscell{\underline{74.55}} & \ourscell{\underline{82.20}} & \ourscell{\textbf{79.05}} & \ourscell{\underline{79.82}} & \ourscell{\textbf{78.99}} & \ourscell{\underline{78.82}} \\
      \midrule
      \multirow{5}{*}{\modelname{Qwen3-8B-it}} 
      & {Mean Pooling} & $\phantom{^{*}}4096\phantom{^{*}}$ & 58.44 & 37.51 & 33.99 & 47.62 & 46.30 & 29.27 & 41.05 & 42.03 \\
      & {PromptEOL}~\cite{prompteol} & $\phantom{^{*}}4096\phantom{^{*}}$ & 50.62 & 66.34 & 55.32 & 68.43 & 75.46 & 72.63 & 64.44 & 64.75 \\
      & {AlignedWVA}~\cite{zhang2026llm} & $\phantom{^{*}}4096\phantom{^{*}}$ & 63.73 & 79.34 & 68.70 & 76.55 & 79.05 & 77.25 & 73.29 & 73.99 \\
      & \ourscell{\textbf{\textsc{DySem} (English Prompt, Ours)}} & \ourscell{$\phantom{^{*}}1066^{*}$} & \ourscell{\underline{74.04}} & \ourscell{\underline{83.18}} & \ourscell{\underline{76.03}} & \ourscell{\underline{82.80}} & \ourscell{\underline{79.31}} & \ourscell{\underline{79.93}} & \ourscell{\underline{76.44}} & \ourscell{\underline{78.82}} \\
      & \ourscell{\textbf{\textsc{DySem} (Language-specific Prompt, Ours)}} & \ourscell{$\phantom{^{*}}728^{*}$} & \ourscell{\textbf{76.43}} & \ourscell{\textbf{84.13}} & \ourscell{\textbf{76.29}} & \ourscell{\textbf{83.81}} & \ourscell{\textbf{81.20}} & \ourscell{\textbf{81.75}} & \ourscell{\textbf{77.50}} & \ourscell{\textbf{80.16}} \\
      \midrule
      \multirow{5}{*}{\modelname{Phi3-mini-it}} 
      & {Mean Pooling} & $\phantom{^{*}}3072\phantom{^{*}}$ & 58.57 & 52.75 & 45.73 & 46.02 & 50.30 & 39.33 & 41.99 & 47.81 \\
      & {PromptEOL}~\cite{prompteol} & $\phantom{^{*}}3072\phantom{^{*}}$ & 57.20 & 70.04 & 60.80 & 71.21 & 73.51 & 70.73 & 63.72 & 66.74 \\
      & {AlignedWVA}~\cite{zhang2026llm} & $\phantom{^{*}}3072\phantom{^{*}}$ & 67.25 & 80.05 & 69.24 & 78.47 & \textbf{80.28} & 77.03 & 72.08 & 74.91 \\
      & \ourscell{\textbf{\textsc{DySem} (English Prompt, Ours)}} & \ourscell{$\phantom{^{*}}987^{*}$} & \ourscell{\underline{71.63}} & \ourscell{\underline{82.20}} & \ourscell{\underline{74.00}} & \ourscell{\underline{81.30}} & \ourscell{76.37} & \ourscell{\underline{77.15}} & \ourscell{\underline{75.22}} & \ourscell{\underline{76.84}} \\
      & \ourscell{\textbf{\textsc{DySem} (Language-Specific Prompt, Ours)}} & \ourscell{$\phantom{^{*}}869^{*}$} & \ourscell{\textbf{72.39}} & \ourscell{\textbf{82.41}} & \ourscell{\textbf{74.02}} & \ourscell{\textbf{82.81}} & \ourscell{\underline{79.98}} & \ourscell{\textbf{81.20}} & \ourscell{\textbf{76.77}} & \ourscell{\textbf{78.51}} \\
      \midrule 
      \multirow{5}{*}{\modelname{Phi3.5-mini-it}} 
      & {Mean Pooling} & $\phantom{^{*}}3072\phantom{^{*}}$ & 53.55 & 54.63 & 46.27 & 51.68 & 53.20 & 43.63 & 44.05 & 49.57 \\
      & {PromptEOL}~\cite{prompteol} & $\phantom{^{*}}3072\phantom{^{*}}$ & 63.30 & 75.42 & 65.57 & 73.81 & 76.04 & 72.85 & 68.33 & 70.76 \\
      & {AlignedWVA}~\cite{zhang2026llm} & $\phantom{^{*}}3072\phantom{^{*}}$ & 66.35 & 78.00 & 68.61 & 78.05 & \textbf{79.47} & 76.82 & 74.26 & 74.51 \\
      & \ourscell{\textbf{\textsc{DySem} (English Prompt, Ours)}} & \ourscell{$\phantom{^{*}}997^{*}$} & \ourscell{\textbf{71.87}} & \ourscell{\underline{81.40}} & \ourscell{\underline{73.49}} & \ourscell{\underline{80.48}} & \ourscell{\underline{77.27}} & \ourscell{\underline{78.10}} & \ourscell{\textbf{77.36}} & \ourscell{\textbf{77.14}} \\
      & \ourscell{\textbf{\textsc{DySem} (Language-Specific Prompt, Ours)}} & \ourscell{$\phantom{^{*}}658^{*}$} & \ourscell{\underline{71.36}} & \ourscell{\textbf{81.41}} & \ourscell{\textbf{73.65}} & \ourscell{\textbf{81.21}} & \ourscell{74.65} & \ourscell{\textbf{78.67}} & \ourscell{\underline{75.76}} & \ourscell{\underline{76.67}} \\

      \bottomrule
    \end{tabular}%
  }
  \caption{Comparison between baselines and our method on instruct models.}
  \label{tab:main_exp_it}
\end{table*}

\section{Experiments}

\subsection{Experimental Settings}

\paragraph{Models.} We use ten LLMs covering both base and instruction-tuned variants from three families: (1) LLaMA~\cite{touvron2023llama, grattafiori2024llama}: LLaMA2-7B, LLaMA2-7B-it, LLaMA3.1-8B, and LLaMA3.1-8B-it; (2) Qwen~\cite{yang2024qwen2,yang2025qwen3}: Qwen2.5-7B, Qwen2.5-7B-it, Qwen3-8B, and Qwen3-8B-it; (3) Phi~\cite{abdin2024phi}: Phi3-mini-it, and Phi3.5-mini-it.

\paragraph{Baselines.}

We compare our method against pre-trained model based methods~\cite{reimers-gurevych-2019-sentence,gao-etal-2021-simcse,jiang-etal-2022-promptbert} and recent LLMs base methods including: \textbf{PromptEOL}~\cite{prompteol}, \textbf{LLM2Vec}~\cite{behnamghader2024llmvec}, \textbf{Token Prepending}~\cite{fu-etal-2025-token}, \textbf{Contrastive Prompting}~\cite{cheng-etal-2025-contrastive}, \textbf{SemPA}~\cite{chen2026sempa}, and \textbf{AlignedWVA}~\cite{zhang2026llm}. See Appendix~\ref{app:baseline} for introduction of these baselines.

\paragraph{Datasets.}
Seven STS datasets are evaluated, STS 2012$\sim$2016 ~\cite{agirre-etal-2012-semeval,agirre-etal-2013-sem,agirre-etal-2014-semeval,agirre-etal-2015-semeval,agirre-etal-2016-semeval}, STS-B~\cite{cer-etal-2017-semeval}, and SICK-R~\cite{marelli-etal-2014-sick}, see Appendix~\ref{app:datasets} for examples. We reported Spearman correlation ($\times 100$).

\paragraph{Implementations.}
For the main results, we use cumulative attention components $A_n^L$ in Eq.~\ref{eq:cum_att_ffn} and top-1024 semantic dimensions in Eq.~\ref{eq:topk} across all models. Both {English prompt} and {language-specific prompt} are used. See Appendix~\ref{app:implementation_details} for detailed settings of the vector selection in Eq.~\ref{eq:vector} and the specific number of languages used.

\subsection{Main Results}

The results are shown in Table~\ref{tab:main_exp_base} and ~\ref{tab:main_exp_it}, respectively. 
\textsc{DySem} achieves the best and second-best average results across all ten LLMs while requiring lower dimensions. Moreover, a paired two-tailed Student's $t$-test over all 70 model-dataset instances confirms that \textsc{DySem} significantly outperforms AlignedWVA ($t=8.1948$, $p=4.40 \times 10^{-12}$), demonstrating its effectiveness.

\paragraph{Base Models.} For LLaMA2-7B base model, \textsc{DySem} achieves the highest average score of 78.98 among all compared methods. For Qwen3-8B, it further exhibits outstanding results of 80.54, surpassing AlignedWVA by 5.75 points and PromptEOL by 11.63 points. We also find that while achieving the best results, the dimensionality required by the model can be as low as 658, indicating that in the Qwen3-8B model, \textsc{DySem} has indeed identified valuable internal components with significant semantic knowledge.

\paragraph{Instruct Models.}
A similar pattern holds on instruct models, where \textsc{DySem} achieves the best average score in all settings and maintains a computational dimensionality of around 1,000 or fewer. Applying our method to LLaMA3.1-8B-it yields an average score of 81.20, with gains of 4.47 points over AlignedWVA and 8.03 points over PromptEOL. The improvement is also substantial on Qwen3-8B-it, where \textsc{DySem} achieves an average score of 80.16 and outperforms AlignedWVA and PromptEOL by 6.17 and 15.41 points, respectively. 
Across instruction-tuned models, the average improvement over \textsc{AlignedWVA} is 4.55 points on instruct models, which is larger than the 3.82 point gain on base models. The reason can be that instruction tuning 
introduces more non-semantic variation inside models and \textsc{DySem} is able to uncover more relevant semantic information to enable better semantic computation.


\section{Analyses and Discussions}
We further analyze the effectiveness of \textsc{DySem}. Below we report the results using representative LLMs, and see Appendix~\ref{app:more_result} for additional ones.

\subsection{Validity of Semantic Components}
\label{subsec:analysis_component_validity}

\begin{figure}[t!]
  \centering
  \includegraphics[width=\columnwidth]{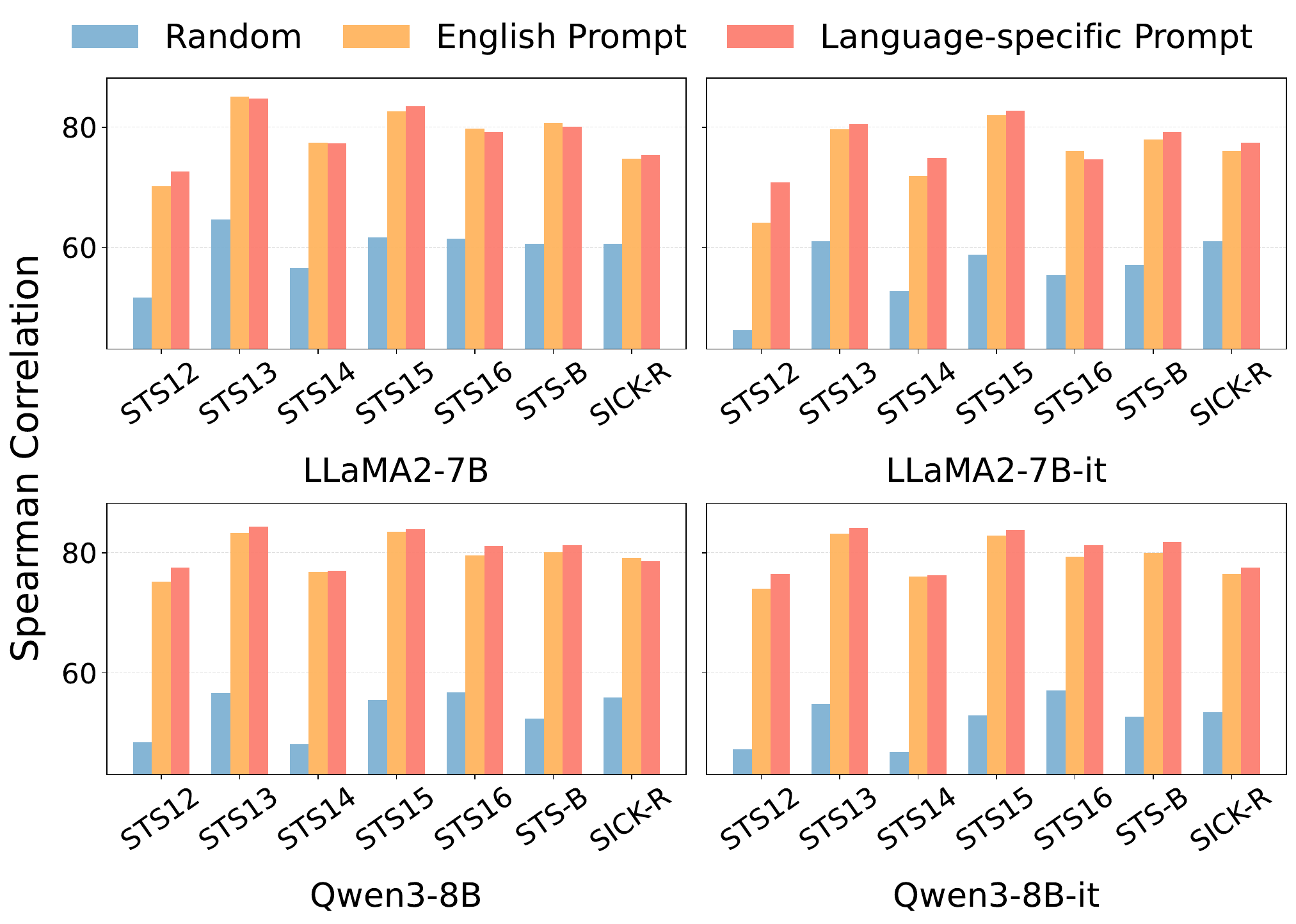}
  \caption{STS performance comparison between random components and semantic components.}
  \label{fig:analyse0_task_bars}
\end{figure}

To verify the effectiveness of semantic components, we compare them with a random component baseline. The random baseline uniformly samples 1024 dimensions from the text representation. Results are shown in Figure~\ref{fig:analyse0_task_bars}.

The results show that semantic components consistently outperform random components across all 70 model-task comparisons, yielding a clear performance gap.
On average, the random baseline scores only 52.70, whereas \textsc{DySem} achieves 78.24 and 79.04 with English and language-specific prompts, respectively. This indicates that the gains of \textsc{DySem} come from identifying meaningful semantic components rather than merely using a smaller subset of features.

\subsection{Semantic Components Selection}
\label{subsec:analysis_component_selection}

\begin{figure}[t!]
  \centering
  \includegraphics[width=\columnwidth]{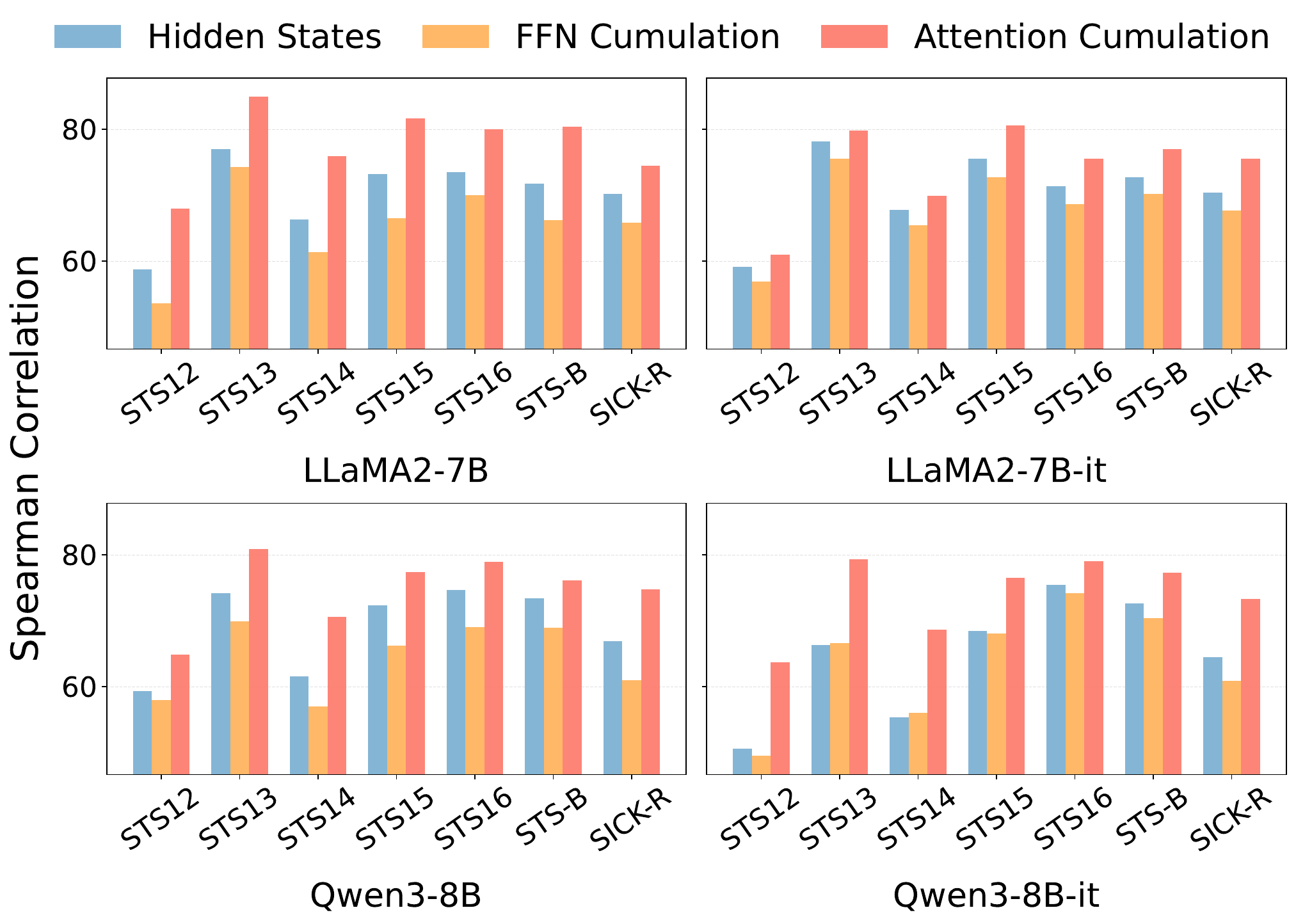}
  \caption{STS performance by selecting different internal components for semantic similarity calculation.}
  \label{fig:analyse1_task_bars}
\end{figure}

\begin{figure}[t!]
  \centering
  \includegraphics[width=\columnwidth]{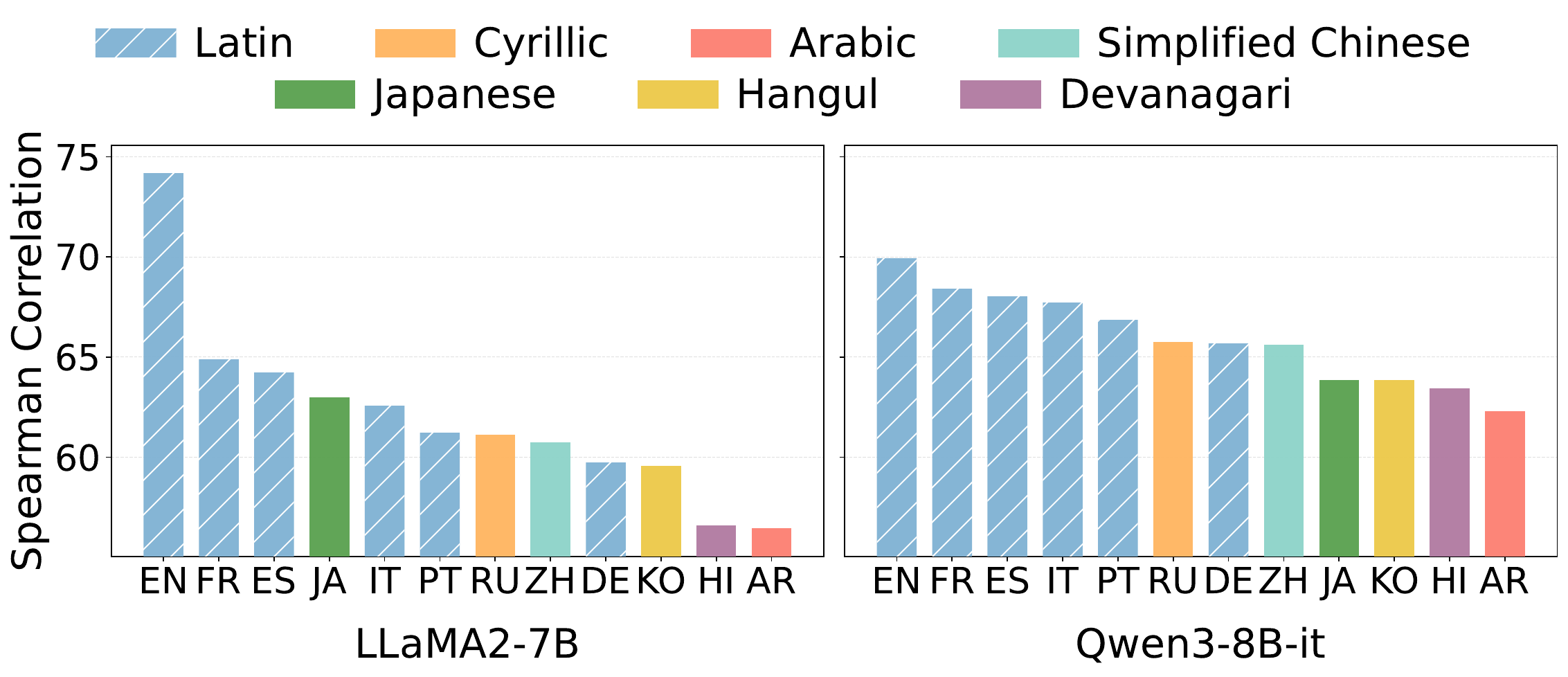}
  \caption{STS performance with texts in different candidate languages based on attention-based components.}
  \label{fig:analyse2_bars}
\end{figure}

To identify the best internal representation for STS calculation, we systematically evaluate three components across ten models: the final-layer hidden states ($h_n^L$), the cumulative FFN outputs ($F_n^L$), and the cumulative attention outputs ($A_n^L$). 

Results are shown in Figure~\ref{fig:analyse1_task_bars}. The results demonstrate that $h_n^L$ is not universally optimal, while $A_n^L$ performs better in most cases. Across 70 comparisons spanning different models and tasks, $A_n^L$ achieves the best performance in 61 cases, while $h_n^L$ takes the lead in 9, and $F_n$ never ranks first. 
Overall, our analyses indicate that \textbf{attention-based representation $A_n^L$ 
is better suited to
semantic similarity than the conventional default hidden states $h_n^L$}. The results reveal that $A_n^L$ could also capture the semantic-level features that can be used for textual similarity computation. 

\subsection{Language Selection Strategies}

To select the language set $\mathcal{L}$ in Eq.~\ref{eq:languageset} for aggregation, we rank candidate languages based on their independent performance and retain the top-$m$ of them. Figure~\ref{fig:analyse2_bars} shows that English
consistently ranks first across models, with French and Spanish usually near the top. 
The reason can be that the model is more thoroughly trained on this type of Latin texts, resulting in a better understanding of them. We also find that the ordering of mid-tier languages varies substantially across models, so that language selection should be model-specific rather than relying on a fixed universal set.

\begin{figure}[t!]
  \centering
  \includegraphics[width=\columnwidth]{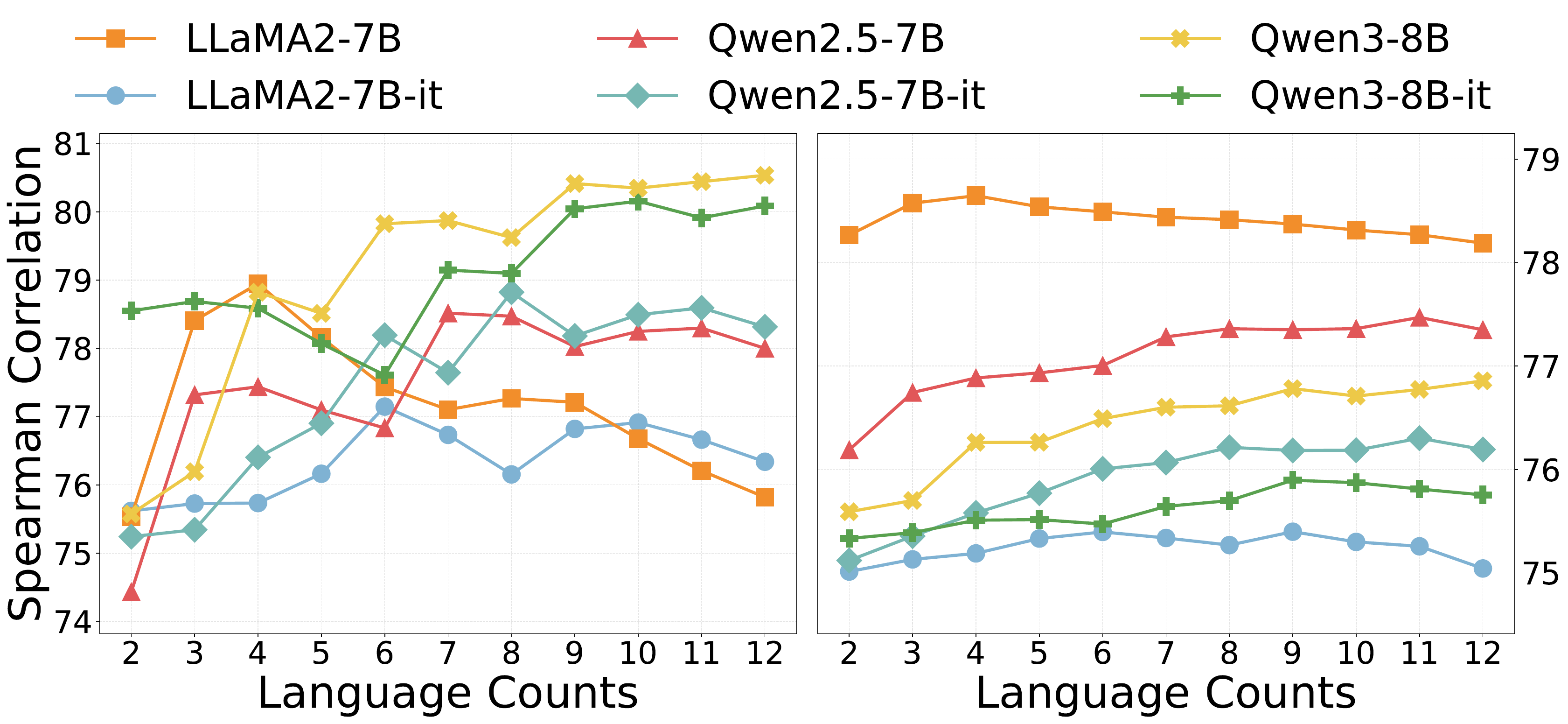}
  \caption{Comparison across different top-$m$ language selections on language-specific prompt with mean vector (left) and source vector (right).}
  \label{fig:analyse2_language-mean_language-source_curves}
\end{figure}

Figure~\ref{fig:analyse2_language-mean_language-source_curves} illustrates how varying the pool size $m$ affects different configurations. Under the language-specific prompt and mean vector, performance is highly sensitive to $m$. Models like LLaMA2-7B peak early at $m=4$, suggesting that incorporating lower-ranked languages introduces semantic noise. Conversely, models with stronger multilingual ability, such as Qwen3-8B, benefit from larger language pools. In contrast, the English prompt and source vector yields flatter performance curves. Together, these trends demonstrate that both careful language selection and the tuning of $m$ play a critical role in determining performance across various models and configurations.

\subsection{Impacts of Prompts and Semantic Vector}

\begin{figure}[t!]
  \centering
  \includegraphics[width=\columnwidth]{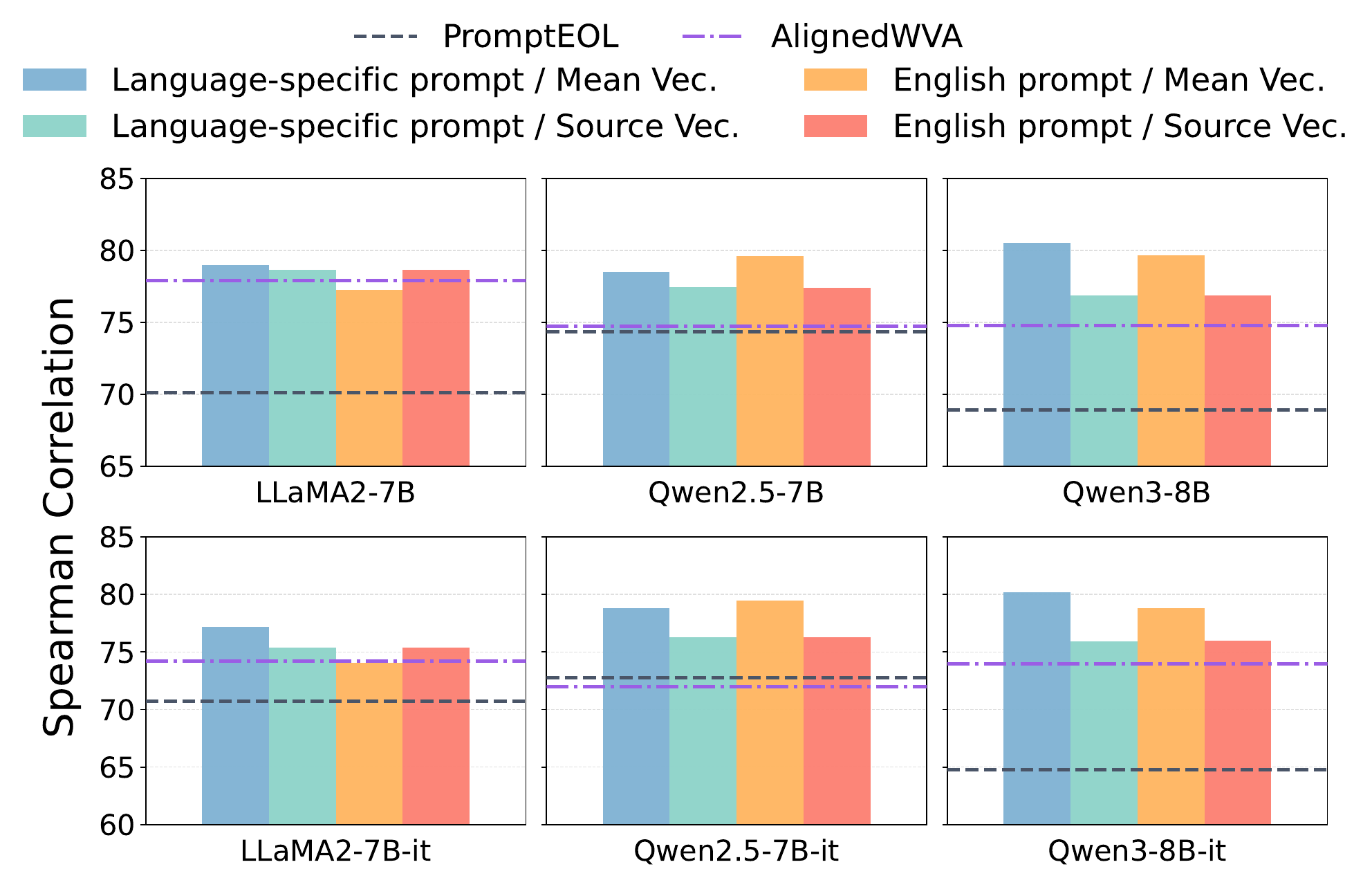}
  \caption{Comparison of English and language-specific prompts under source and mean vectors settings.}
  \label{fig:analyse3_bars}
\end{figure}

We analyze the impact of English and language-specific prompts, alongside two semantic vector constructions, namely source and mean vectors. As illustrated in Figure~\ref{fig:analyse3_bars}, mean vector generally outperform source vector, especially under language-specific prompts, indicating that vector construction is the primary driver of performance.

The combination of a language-specific prompt and a mean vector achieves the highest average score on seven of the ten models. Notably, the choice of prompt language strongly interacts with vector construction. 
With the mean vector, language-specific prompts improve performance by 1.24 points on average, whereas with the source vector, prompt language differences are negligible, within 0.39 points.
This suggests that language-specific prompts are most effective when translated-language representations are explicitly aggregated into the final semantic vector. Overall, our method significantly outperforms baselines across different settings, demonstrating that we indeed extracts more effective semantic features.

\begin{figure}[t!]
  \centering
  \includegraphics[width=\columnwidth]{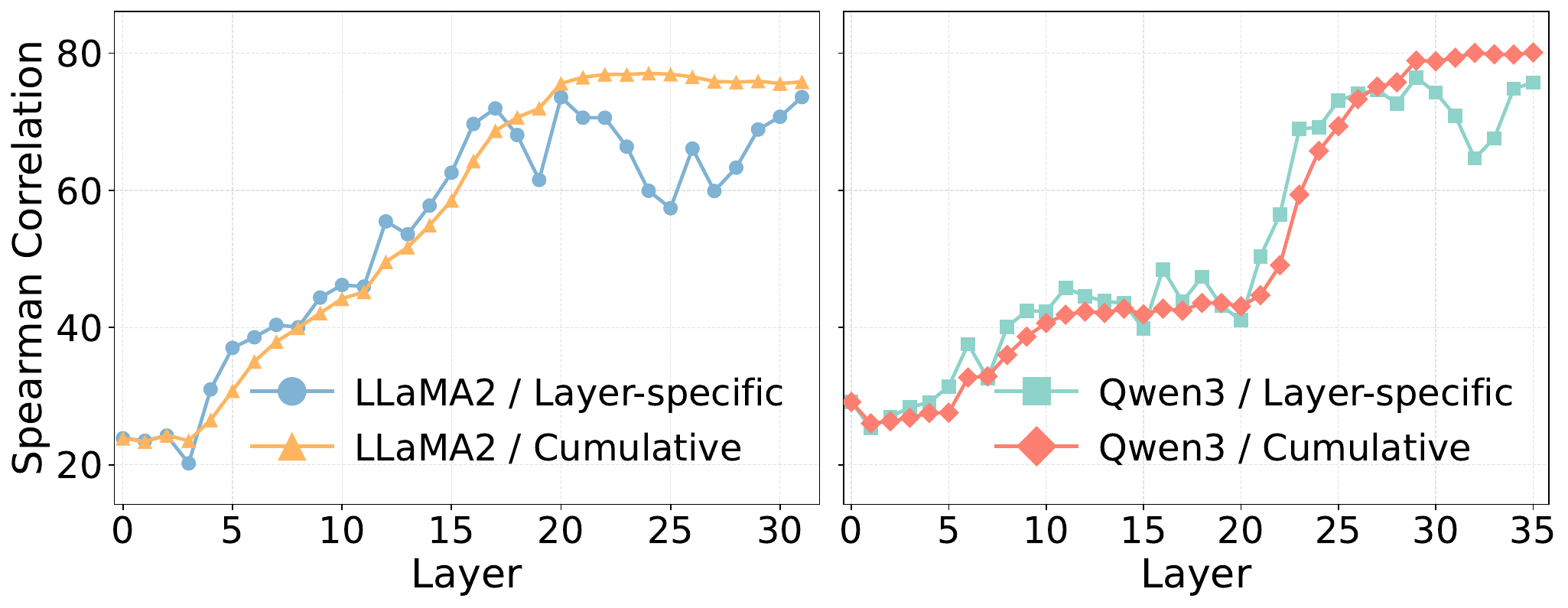}
  \caption{Comparison between layer-wise attention and cumulative attention aggregation strategies on LLaMA2-7B (left) and Qwen3-8B-it (right). }
  \label{fig:analyse4_curves}
\end{figure}

\subsection{Comparing Two Attention Components}

We further compare two ways to construct attention-based semantic components: (1) \textbf{layer-specific attention} $a_n^l$, which uses only the output of layer $l$, and (2) \textbf{cumulative attention} $A_n^l$, which accumulates outputs from the bottom layer up to layer $l$.  The results with language-specific prompt and mean vector settings are shown in Figure~\ref{fig:analyse4_curves}.

Overall, the cumulative strategy outperforms the layer-specific strategy with more stable trends, especially in the middle-to-upper layers.
For instance, peak performance over layer-specific attention rise
from 76.43 to 80.09 on Qwen3-8B-it. These findings indicate that semantic information is distributed across layers rather than concentrated in one. 
We therefore use the final cumulative attention $A_n^L$ defined in \S\ref{subsec:decoder_components} to better capture semantics.

\begin{figure}[t!]
  \centering
  \includegraphics[width=\columnwidth]{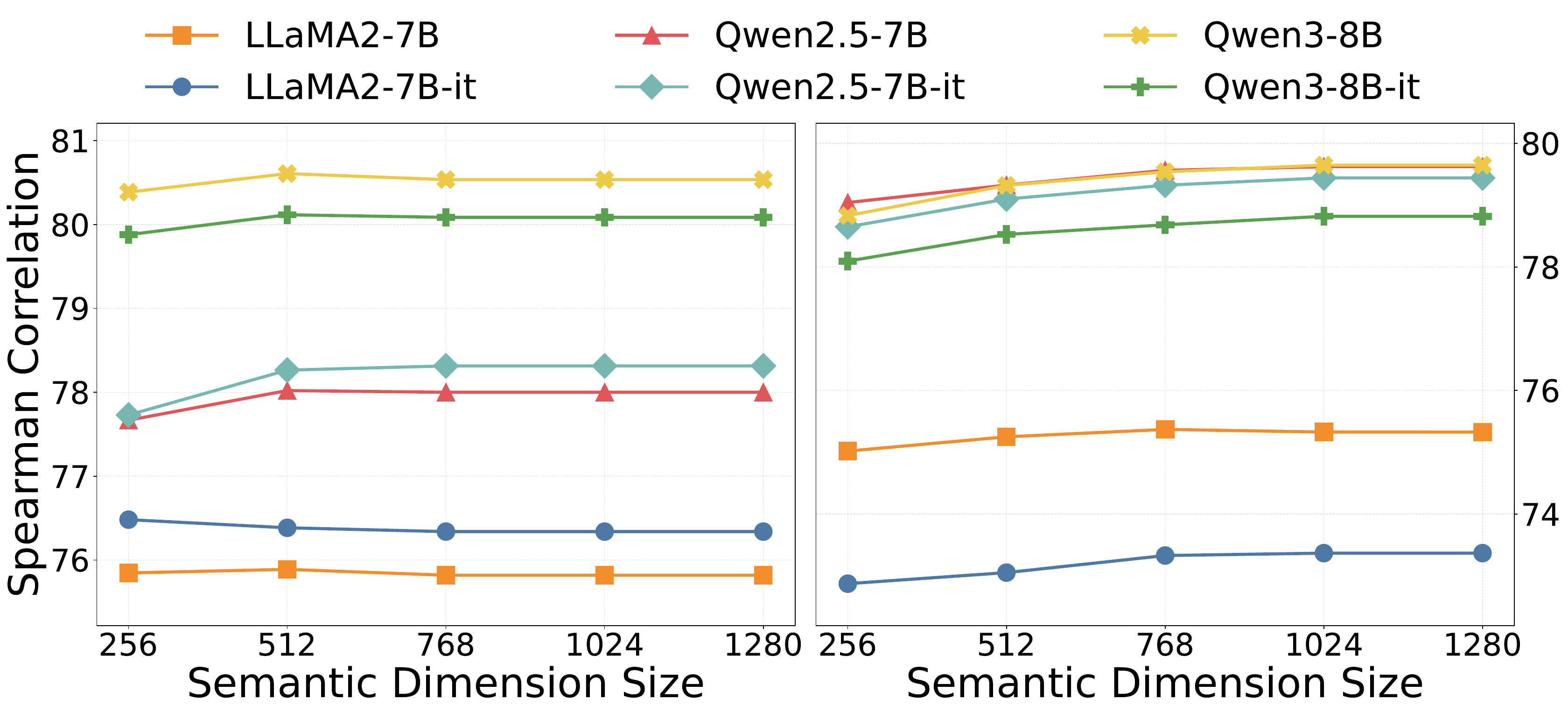}
  \caption{Analysis of top-$k$ semantic dimension selection on language-specific prompt (left) and English prompt (right) with mean vector .}
  \label{fig:analyse5_language-mean_english-mean_curves}
\end{figure}

\subsection{Impacts of Semantic Dimension Sizes}
\label{subsec:analysis_dimension_size}

We compare top-$k$ dimension selections for $k \in \{256, 512, 768, 1024, 1280\}$ in Eq.~\ref{eq:topk}. As illustrated in Figure~\ref{fig:analyse5_language-mean_english-mean_curves}, performance patterns are primarily dictated by the prompt language. Under language-specific prompt settings, performance improves sharply up to $k=512$ and then saturates. 


By contrast, English prompt benefit from retaining significantly more dimensions, exhibiting an almost monotonic improvement up to $k=1280$. 
These results indicate that semantic computation can be performed in lower dimensions, without necessarily using full dimensions of the hidden states (e.g., the typical size of 4096 for LLMs).

\subsection{Analysis of Computational Overhead}

We evaluate the computational overhead of \textsc{DySem} using language-specific prompts, mean-vector aggregation, and four languages. As shown in Table~\ref{tab:computational_overhead}, \textsc{DySem} requires $4.21\times$ and $3.45\times$ the runtime of AlignedWVA on LLaMA2-7B and Qwen3-8B, respectively. This additional cost mainly arises from extracting and evaluating four language-specific representations. Nevertheless, the overhead enables \textsc{DySem} to identify more compact and semantically informative dimensions, providing a practical trade-off between computational cost and representation quality.

\subsection{Discussion on Further Applications}

\paragraph{Multi-modal Applications.}
Although our dynamic semantic extraction framework is currently evaluated on text representation based on multilingual consensus, its core principle of identifying invariant dimensions across varied surface expressions can be naturally extended to broader vision domains. For example, treating images with operations like rotations or 3D point clouds with different camera angles as parallel ``translations'' can also used for extract shared semantic components.

\paragraph{Practical Deployments.}
Our method can also be used for practical deployment such as text indexing and retrieval. By caching just the semantic dimension set $\mathcal{S}(y)$ alongside the representation vector for each database sample, the joint semantic set $\mathcal{U}(x, y)$ for a new query $x$ can be dynamically constructed on the fly. This allows for direct similarity computation without the computational overhead of re-encoding existing database samples. 
Ultimately, dynamic dimension selection serves as a lightweight and general processing step before similarity calculation and text retrieval.

\begin{table}[t]
    \centering
    \caption{Runtime comparison between AlignedWVA and \textsc{DySem}.}
    \label{tab:computational_overhead}
    \footnotesize
    \setlength{\tabcolsep}{3pt}
    \renewcommand{\arraystretch}{1}
    \begin{tabular}{llcc}
        \toprule
        Model & Method & Runtime (s) & Relative Runtime \\
        \midrule
        LLaMA2-7B & AlignedWVA     & 172.25 & $1.00\times$ \\
        LLaMA2-7B & \textsc{DySem} & 725.45 & $4.21\times$ \\
        \midrule
        Qwen3-8B  & AlignedWVA     & 169.49 & $1.00\times$ \\
        Qwen3-8B  & \textsc{DySem} & 585.33 & $3.45\times$ \\
        \bottomrule
    \end{tabular}
\end{table}

\section{Related Work}

\paragraph{Semantic Representation with LLMs.}
Decoder-based LLMs originally target generation and recent works have leveraged them for extracting semantic representations. 
PromptEOL~\cite{prompteol} and MetaEOL~\cite{metaeol} both extract text semantics solely through prompt templates, with the latter further introducing multi-perspective prompts to enrich representations.
\citet{fu-etal-2025-token} and \citet{cheng-etal-2025-contrastive} improve prompt-based LLM sentence embeddings through training-free inference-stage modifications, respectively by prepending decoded sentence embeddings to mitigate causal-attention limitations and by contrastively steering hidden states to emphasize core semantics over non-essential information.
Recent embedding models, including NV-Embed~\cite{lee2025nvembed}, Qwen3-Embedding~\cite{zhang2025qwen3embeddingadvancingtext}, and Gemini Embedding~\cite{lee2025gemini}, often rely on proprietary training data.
These works rely on fixed, full-dimensional hidden states for text representation. Our work addresses these issues by directly mining internal LLM modules to extract dynamic semantics in a training-free manner.

\paragraph{Semantic Knowledge of Models.}
Studies have shown that linguistic information is hierarchically organized across model layers: shallow layers predominantly capture surface-level and syntactic features, while deeper intermediate layers encode richer semantic and conceptual content~\cite{jawahar-etal-2019-bert, ethayarajh-2019-contextual, liu-etal-2019-linguistic}. 
Recent evidence further suggests that intermediate representations can provide stronger feature quality than final-layer embeddings across a wide range of text-embedding tasks, indicating that semantic information is not confined to the output end of the model~\cite{skean2025layer}.
Recent work by \citet{zhang2026llm} shows that attention value vectors also encode some semantic knowledge. 
Inspired by them, we further investigate internal dynamic semantic component via multilingual consensus and novelly calculate semantic relationships with flexible dimensions, an aspect unexplored in prior work.

\section{Conclusion}

We propose \textsc{DySem}, a training-free framework for better calculating semantic textual similarity beyond the vanilla way of text representation which uses fixed final-layer hidden states and static full-dimensional vectors. \textsc{DySem} uncovers dynamic sample-specific semantic component via multilingual consensus, and uses flexible dimensions from semantic sets for similarity calculation. Extensive experiments demonstrate that \textsc{DySem} achieves superior performance across standard STS benchmarks, consistently outperforming a variety of base and instruction-tuned LLMs while maintaining lower dimensions. We hope that this work will provide new directions for better representing texts and evaluating their relationships with LLMs.

\section*{Limitations}

While our method demonstrates strong performance, extracting semantic dimensions introduces an $O(m)$ inference overhead due to the required $m$ translations and independent LLM forward passes. However, this cost is primarily concentrated in the offline database building phase; online queries remain highly efficient by dynamically constructing the joint semantic set $\mathcal{U}(x, y)$ from cached dimensions. Nonetheless, the initial encoding cost poses a challenge for massive-scale, cold-start applications compared to traditional single-pass encoders.

Additionally, our work thoroughly validates STS and conducts extensive experiments. Although STS directly measures text representation quality, future work can apply it on a broader suite of representation-heavy downstream tasks, such as large-scale information retrieval, text clustering, and zero-shot classification. Moreover, our framework primarily focuses on semantic computation between two texts, rather than high-quality embedding of a single text; therefore, we focus on extract semantics in a training-free way of LLMs and do not compare it with pure embedding models.

\section*{Acknowledgments}

This work was supported in part by ICFCRT (W2441020), Guangdong Basic and Applied Basic Research Foundation (2026A1515011358), Youth S\&T Talent Support Programme of Guangdong Provincial Association for Science and Technology (SKXRC2026862), Shenzhen Natural Science Foundation (JCYJ20250604181610014), Intelligent Computing Center of Shenzhen University, and Scientific Development Fund from Guangdong Provincial Key Laboratory of Visual Media and Multidimensional Intelligence.

\bibliography{custom}

\appendix
\section{Details of Baselines}
\label{app:baseline}

We compare our method against following baselines: \textbf{PromptEOL} ~\cite{prompteol}, a standard hidden-state-based prompt approach; \textbf{Token Prepending} ~\cite{fu-etal-2025-token}, a training-free method that prepends tokens to mitigate causal-attention bias; \textbf{Contrastive Prompting} ~\cite{cheng-etal-2025-contrastive}, an inference-time method that steers semantic encoding with contrastive prompts; \textbf{LLM2Vec} ~\cite{behnamghader2024llmvec}, an unsupervised method that converts decoder-only LLMs into strong text encoders by enabling bidirectional attention, masked next-token prediction, and contrastive learning; \textbf{SemPA} ~\cite{chen2026sempa}, a semantic preference alignment method that improves sentence representations with sentence-level DPO; \textbf{AlignedWVA} ~\cite{zhang2026llm}, a training-free LLM-based embedding method that leverages attention outputs vectors for sentence representation and serves as our main attention-based baseline. Moreover, we compare with classical BERT-based sentence embedding baselines: \textbf{Sentence-BERT}~\cite{reimers-gurevych-2019-sentence}, which employs Siamese and triplet network structures to derive semantically meaningful sentence embeddings; \textbf{SimCSE}~\cite{gao-etal-2021-simcse}, a simple contrastive learning framework that learns sentence embeddings using dropout-based augmentation in the unsupervised setting and NLI supervision in the supervised setting; and \textbf{PromptBERT}~\cite{jiang-etal-2022-promptbert}, a prompt-based contrastive learning method that improves BERT sentence embeddings by reformulating sentence representation with prompts and template denoising.

\section{Example of Datasets}
\label{app:datasets}

Table~\ref{tab:data_example} provides representative sentence pairs from the evaluation datasets, with dataset sizes shown in parentheses.

\begin{table}[ht]
  \centering
  \large
  \renewcommand{\arraystretch}{1.2}
  
  \resizebox{\linewidth}{!}{
    \begin{tabular}{lp{10cm}}
      \toprule
      \textbf{Type} & \textbf{Content} \\
      
      \midrule
      
      \rowcolor{gray!20}
      \multicolumn{2}{c}{\textbf{STS12 (3108)}} \\ 
      \textbf{Sentence1} & the problem likely will mean corrective changes before the shuttle fleet starts flying again . \\
      \textbf{Sentence2} & he said the problem needs to be corrected before the space shuttle fleet is cleared to fly again . \\
      \textbf{Score}   & \boxed{4.4} \\
      \textbf{Link}    & \href{https://huggingface.co/datasets/mteb/sts12-sts}{https://huggingface.co/datasets/mteb/sts12-sts} \\

      \midrule
      \rowcolor{gray!20}
      \multicolumn{2}{c}{\textbf{STS13 (1500)}} \\ 
      \textbf{Sentence1} & this frame contains words that describe an item 's static position on a scale with respect to some property variable . \\
      \textbf{Sentence2} & lacking in specific resources , qualities or substances ; \\
      \textbf{Score}   & \boxed{0.4} \\
      \textbf{Link}    & \href{https://huggingface.co/datasets/mteb/sts13-sts}{https://huggingface.co/datasets/mteb/sts13-sts} \\

      \midrule
      \rowcolor{gray!20}
      \multicolumn{2}{c}{\textbf{STS14 (3750)}} \\ 
      \textbf{Sentence1} & how about some testimonies from real health experts ? \\
      \textbf{Sentence2} & also , who 's to say there aren 't testimonies from " real " health experts on there ? \\
      \textbf{Score}   & \boxed{3} \\
      \textbf{Link}    & \href{https://huggingface.co/datasets/mteb/sts14-sts}{https://huggingface.co/datasets/mteb/sts14-sts} \\
      
      \midrule
      \rowcolor{gray!20}
      \multicolumn{2}{c}{\textbf{STS15 (3000)}} \\ 
      \textbf{Sentence1} & you 'll need to check the particular policies of each publisher to see what is allowed and what is not allowed . \\
      \textbf{Sentence2} & if you need to publish the book and you have found one publisher that allows it . \\
      \textbf{Score}   & \boxed{3} \\
      \textbf{Link}    & \href{https://huggingface.co/datasets/mteb/sts15-sts}{https://huggingface.co/datasets/mteb/sts15-sts} \\

      \midrule
      \rowcolor{gray!20}
      \multicolumn{2}{c}{\textbf{STS16 (3108)}} \\ 
      \textbf{Sentence1} & what is the best way to store fresh berries ? \\
      \textbf{Sentence2} & what is the best way to fix this garage floor ? \\
      \textbf{Score}   & \boxed{0} \\
      \textbf{Link}    & \href{https://huggingface.co/datasets/mteb/sts16-sts}{https://huggingface.co/datasets/mteb/sts16-sts} \\

      \midrule
      \rowcolor{gray!20}
      \multicolumn{2}{c}{\textbf{STSBenchmark (1379)}} \\ 
      \textbf{Sentence1} & A girl is styling her hair. \\
      \textbf{Sentence2} & A girl is brushing her hair. \\
      \textbf{Score}   & \boxed{2.5} \\
      \textbf{Link}    & \href{https://huggingface.co/datasets/mteb/stsbenchmark-sts}{https://huggingface.co/datasets/mteb/stsbenchmark-sts} \\

      \midrule
      \rowcolor{gray!20}
      \multicolumn{2}{c}{\textbf{STS-R (9927)}} \\ 
      \textbf{Sentence1} & A group of kids is playing in a yard and an old man is standing in the background \\
      \textbf{Sentence2} & A group of boys in a yard is playing and a man is standing in the background \\
      \textbf{Score}   & \boxed{4.5} \\
      \textbf{Link}    & \href{https://huggingface.co/datasets/mteb/sickr-sts}{https://huggingface.co/datasets/mteb/sickr-sts} \\
      \bottomrule
    \end{tabular}
  }
  \caption{Examples of sentence pairs and similarity scores from the STS evaluation datasets.}
  \label{tab:data_example}
\end{table}

\section{Implementation Details}
\label{app:implementation_details}

\begin{table}[ht]
\centering
\small
\resizebox{1\columnwidth}{!}{%
    \begin{tabular}{lccc}
    \toprule
    \textbf{Models \& Links} & \shortstack{\textbf{Prompt Setting}} & \shortstack{\textbf{Semantic Vector}} & \shortstack{\textbf{Language Counts}} \\
    \midrule
    \multirow{2.5}{*}{\href{https://huggingface.co/meta-llama/Llama-2-7b-hf}{\textsc{LLaMA2-7B}}} & English & Source & 9 \\
    \cmidrule{2-4}
    & Language-Specific & Mean & 4 \\
    \midrule
    \multirow{2.5}{*}{\href{https://huggingface.co/meta-llama/Llama-3.1-8B}{\textsc{LLaMA3.1-8B}}} & English & Source & 11 \\
    \cmidrule{2-4}
    & Language-Specific & Mean & 6 \\
    \midrule
    \multirow{2.5}{*}{\href{https://huggingface.co/Qwen/Qwen2.5-7B}{\textsc{Qwen2.5-7B}}} & English & Mean & 11 \\
    \cmidrule{2-4}
    & Language-Specific & Mean & 7 \\
    \midrule
    \multirow{2.5}{*}{\href{https://huggingface.co/Qwen/Qwen3-8B-Base}{\textsc{Qwen3-8B}}} & English & Mean & 12 \\
    \cmidrule{2-4}
    & Language-Specific & Mean & 12 \\
    \midrule
    \multirow{2.5}{*}{\href{https://huggingface.co/meta-llama/Llama-2-7b-chat}{\textsc{LLaMA2-7B-it}}} & English & Source & 12 \\
    \cmidrule{2-4}
    & Language-Specific & Mean & 6 \\
    \midrule
    \multirow{2.5}{*}{\href{https://huggingface.co/meta-llama/Llama-3.1-8B-Instruct}{\textsc{LLaMA3.1-8B-it}}} & English & Mean & 12 \\
    \cmidrule{2-4}
    & Language-Specific & Mean & 10 \\
    \midrule
    \multirow{2.5}{*}{\href{https://huggingface.co/Qwen/Qwen2.5-7B-Instruct}{\textsc{Qwen2.5-7B-it}}} & English & Mean & 11 \\
    \cmidrule{2-4}
    & Language-Specific & Mean & 8 \\
    \midrule
    \multirow{2.5}{*}{\href{https://huggingface.co/Qwen/Qwen3-8B}{\textsc{Qwen3-8B-it}}} & English & Mean & 12 \\
    \cmidrule{2-4}
    & Language-Specific & Mean & 10 \\
    \midrule
    \multirow{2.5}{*}{\href{https://huggingface.co/microsoft/Phi-3-mini-128k-instruct}{\textsc{Phi3-mini-it}}} & English & Mean & 8 \\
    \cmidrule{2-4}
    & Language-Specific & Mean & 4 \\
    \midrule
    \multirow{2.5}{*}{\href{https://huggingface.co/microsoft/Phi-3.5-mini-instruct}{\textsc{Phi3.5-mini-it}}} & English & Mean & 10 \\
    \cmidrule{2-4}
    & Language-Specific & Mean & 8 \\
    \bottomrule
    \end{tabular}%
    }
    \caption{Detailed parameter settings for the main experiments across different models, including the chosen prompt setting, semantic vector construction method, and language counts.}
    \label{tab:main_exp_detail_setting}
\end{table}

For the main experiments, our method uses a fixed Semantic Component $A_n^L$ and a selection budget of the top-1024 semantic dimensions across all models. The remaining configurations, namely the prompt setting, semantic vector, and the number of languages, are detailed as follows. Specifically, the English prompt applies a shared English template across all translations, whereas the language-specific prompt tailors the template to the native language of each target translation. For semantic vector construction, source denotes using the representation from the original input text, while mean denotes averaging the representations across the selected languages. The language count indicates how many languages are utilized for semantic dimension selection. As shown in the Table~\ref{tab:main_exp_detail_setting}, the optimal configuration varies by model, suggesting that different model families and instruction-tuning approaches exhibit distinct degrees of multilingual representation stability. Generally, the combination of language-specific prompt and mean semantic vectors is frequently selected, indicating that aggregating multilingual representations helps identify more robust semantic dimensions.

\section{Additional Experimental Results}
\label{app:more_result}

\subsection{Validity of Semantic Components}
\label{app:more_result_analyse0}

\begin{figure*}[t!]
  \centering
  \includegraphics[width=\textwidth]{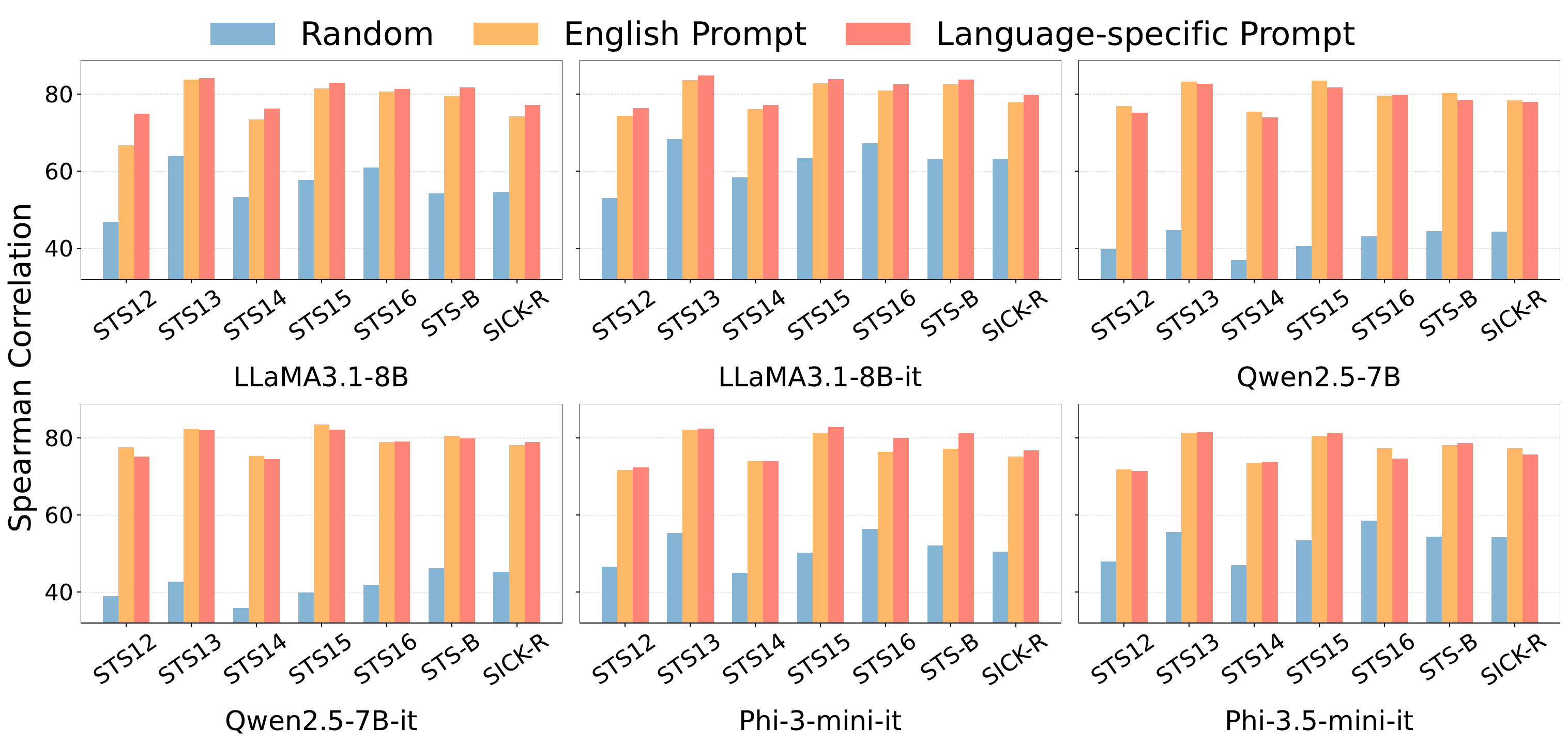}
  \caption{STS performance comparison between random components and semantic components on the remaining models.}
  \label{fig:analyse0_task_bars_remaining_models}
\end{figure*}

Figure~\ref{fig:analyse0_task_bars_remaining_models} shows the complementary results on the remaining six models. The same pattern holds consistently across these models: prompt-selected semantic components substantially outperform random components in all evaluated settings. For example, on Qwen3-8B, language-specific selection improves the average score from 53.41 to 80.53; on LLaMA2-7B-it, it improves the score from 56.02 to 77.16. The gap is especially pronounced for the Qwen2.5 family, where random components obtain an average score of around 42, whereas semantic components reach nearly 79. Similar improvements are also observed for the LLaMA3.1 and Phi models. These results further confirm that the selected components capture task-relevant semantic information, rather than merely benefiting from random feature sparsification.

\subsection{Semantic Components Selection}
\label{app:more_result_analyse1}

\begin{figure*}[ht]
  \centering
  \includegraphics[width=1\textwidth]{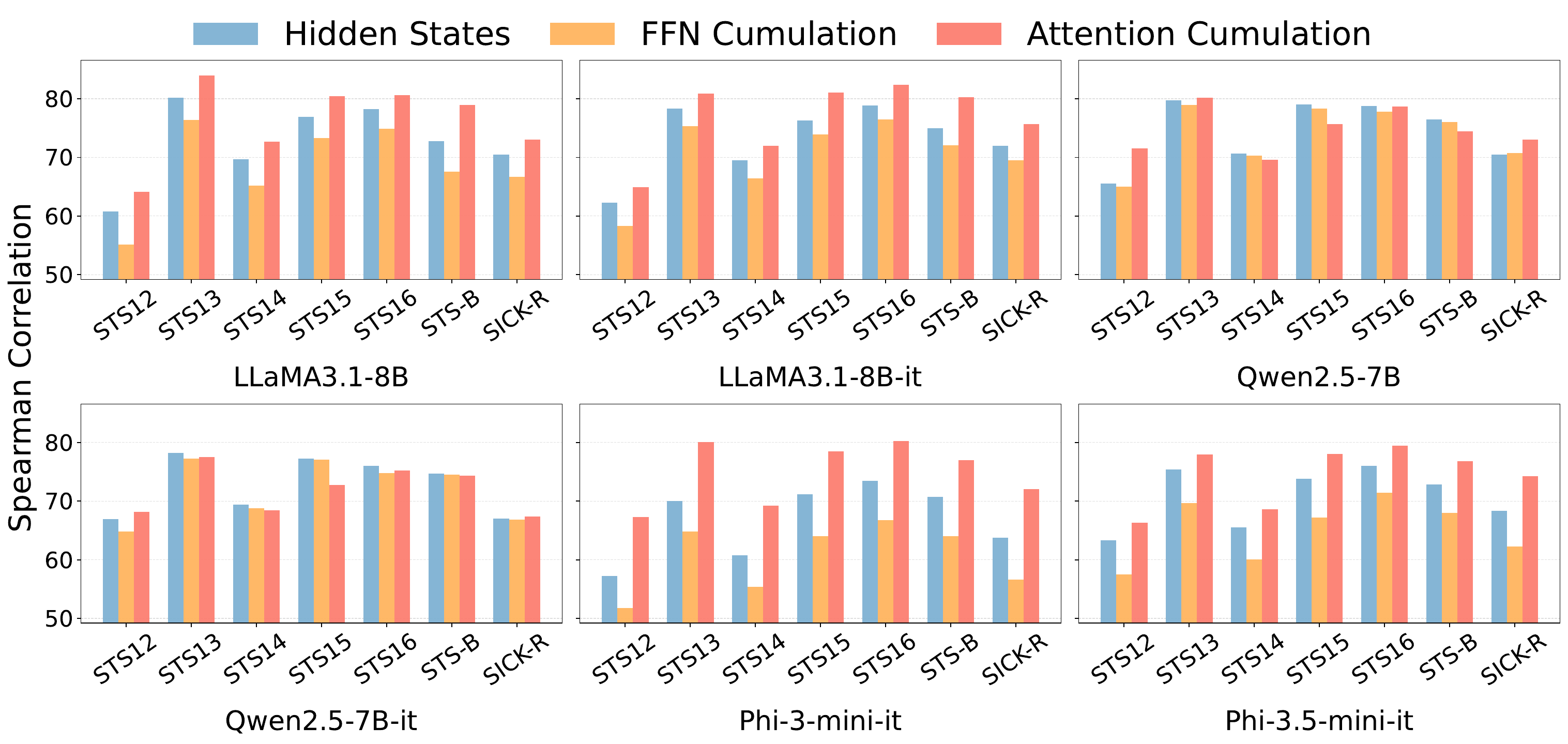}
  \caption{Task-level semantic similarity performance of different internal representations on remaining models.}
  \label{fig:analyse1_task_bars_remaining_models}
\end{figure*}

Figure~\ref{fig:analyse1_task_bars_remaining_models} provides the complementary component-selection results for the remaining six models. The overall trend is consistent with the main analysis: the cumulative attention outputs ($A_n^L$) are usually the strongest representation for text semantic similarity. The advantage of $A_n^L$ is especially clear for LLaMA3.1-8B, LLaMA3.1-8B-it, Phi3-mini-it, and Phi3.5-mini-it, where it outperforms the alternatives on every evaluated STS task. The only substantial exceptions come from the Qwen2.5 family, where $h_n^L$ is competitive and occasionally better on tasks such as STS14, STS15, STS16, and STS-B. Even in these cases, $F_n^L$ remains consistently weaker. This reinforces the conclusion that $A_n^L$ is the most reliable representation for exposing semantic similarity information.

\begin{figure*}[ht]
  \centering
  \includegraphics[width=1\textwidth]{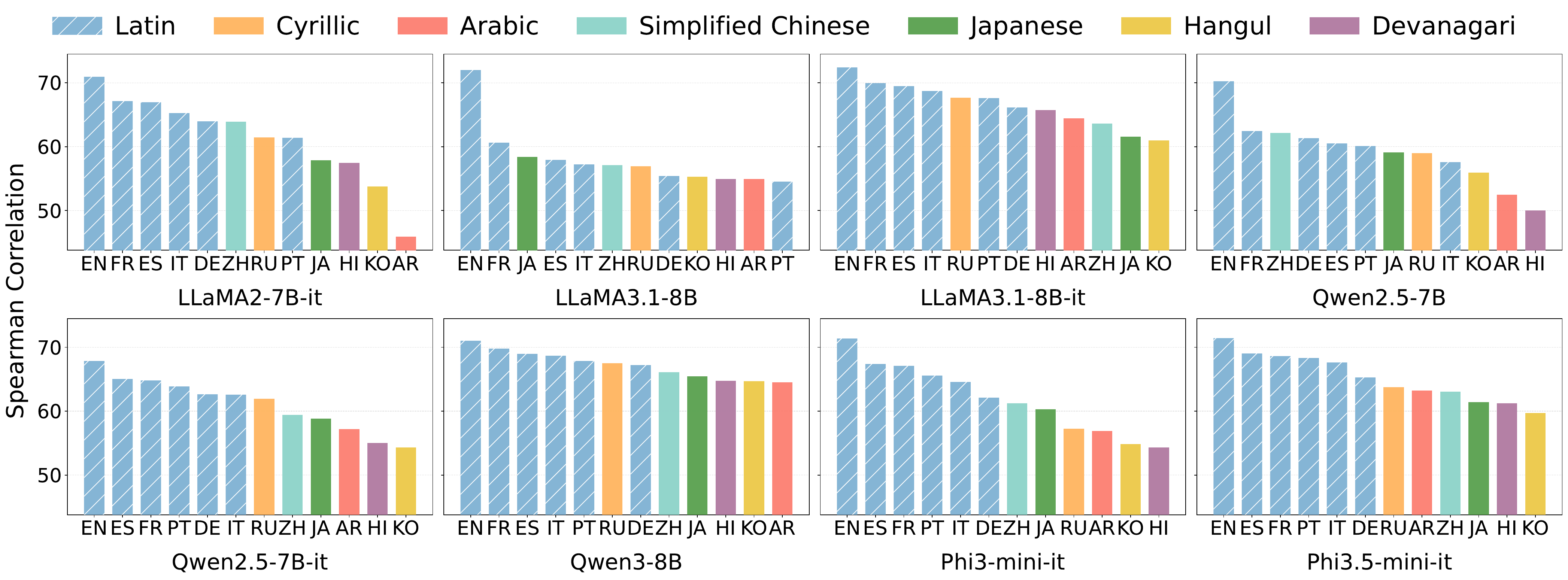}
  \caption{Language-wise spearman correlation of cumulative attention on remaining models.}
  \label{fig:analyse2_other_models_bars}
\end{figure*}

\subsection{Language Selection Strategies}
\label{app:more_result_analyse2}

The appendix provides the full set of language-selection results omitted from the main text. Figure~\ref{fig:analyse2_other_models_bars} extends the language-wise ranking analysis to the remaining models. The overall pattern is consistent with Figure~\ref{fig:analyse2_bars}: English remains the strongest individual language in all cases, and French and Spanish are usually placed near the top. At the same time, the middle-ranked languages differ noticeably across models. For example, Qwen2.5-7B ranks Chinese and German relatively high, LLaMA3.1-8B assigns a high rank to Japanese, while the Phi models favor Portuguese and Italian among the top candidates. This confirms that the ranking is not universal, and that the language pool should be selected separately for each model.

\begin{figure}[ht]
  \centering
  \includegraphics[width=1\columnwidth]{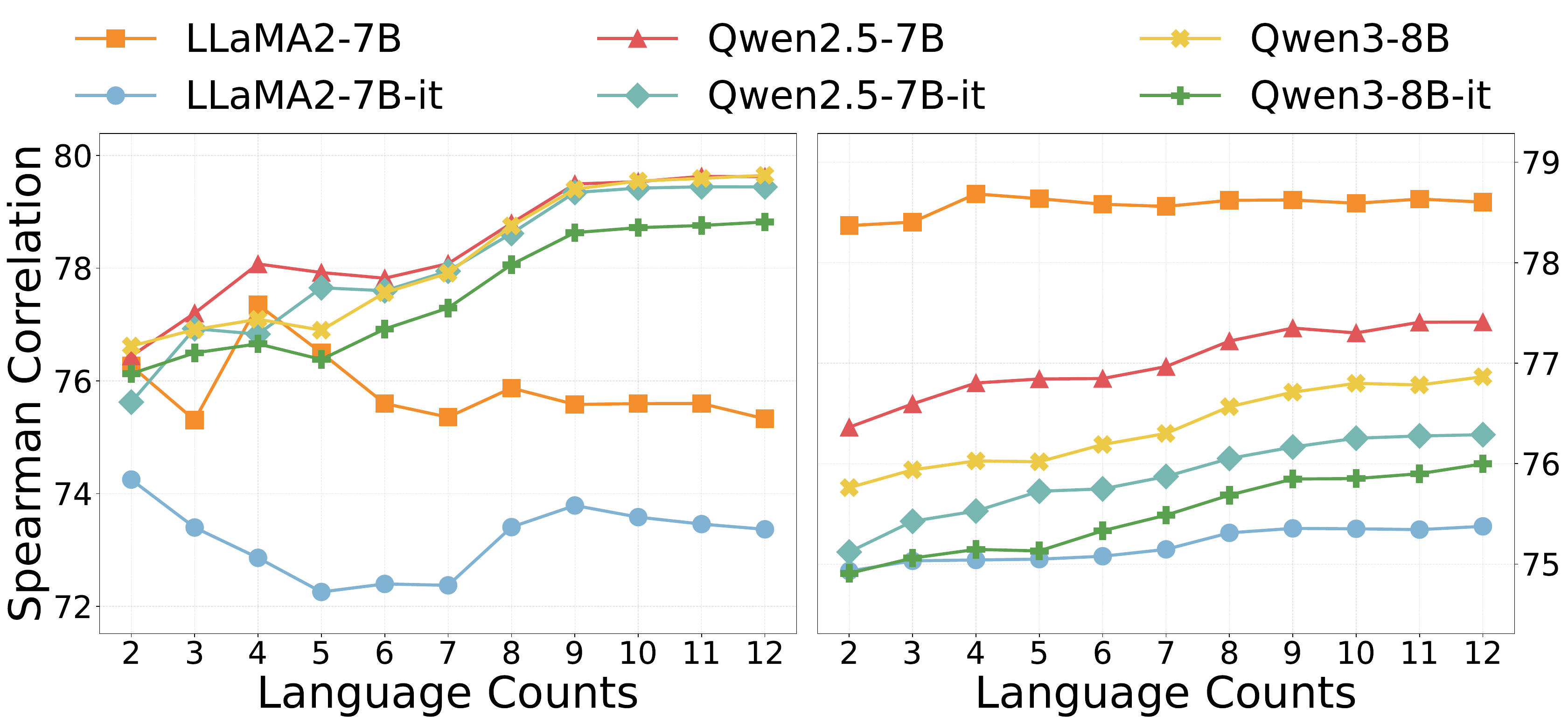}
  \caption{Analysis of language selection strategies on English prompt with mean vector (left) and source vector (right). Comparison across different top-$m$ language selections.}
  \label{fig:analyse2_english-mean_english-source_curves}
\end{figure}

Figure~\ref{fig:analyse2_english-mean_english-source_curves} reports the two complementary configurations for the main models. With the English prompt and mean vector, performance remains sensitive to the number of selected languages: LLaMA2 models reach their best scores with small pools, while Qwen models generally benefit from larger pools and often peak around $m=11$ or $m=12$. In contrast, the language-specific prompt with source vector produces much flatter curves. This suggests that source-vector aggregation is more robust to the inclusion of additional languages, whereas mean-vector aggregation can either amplify useful multilingual signals or introduce noise depending on the model.

\begin{figure}[ht]
  \centering
  \includegraphics[width=1\columnwidth]{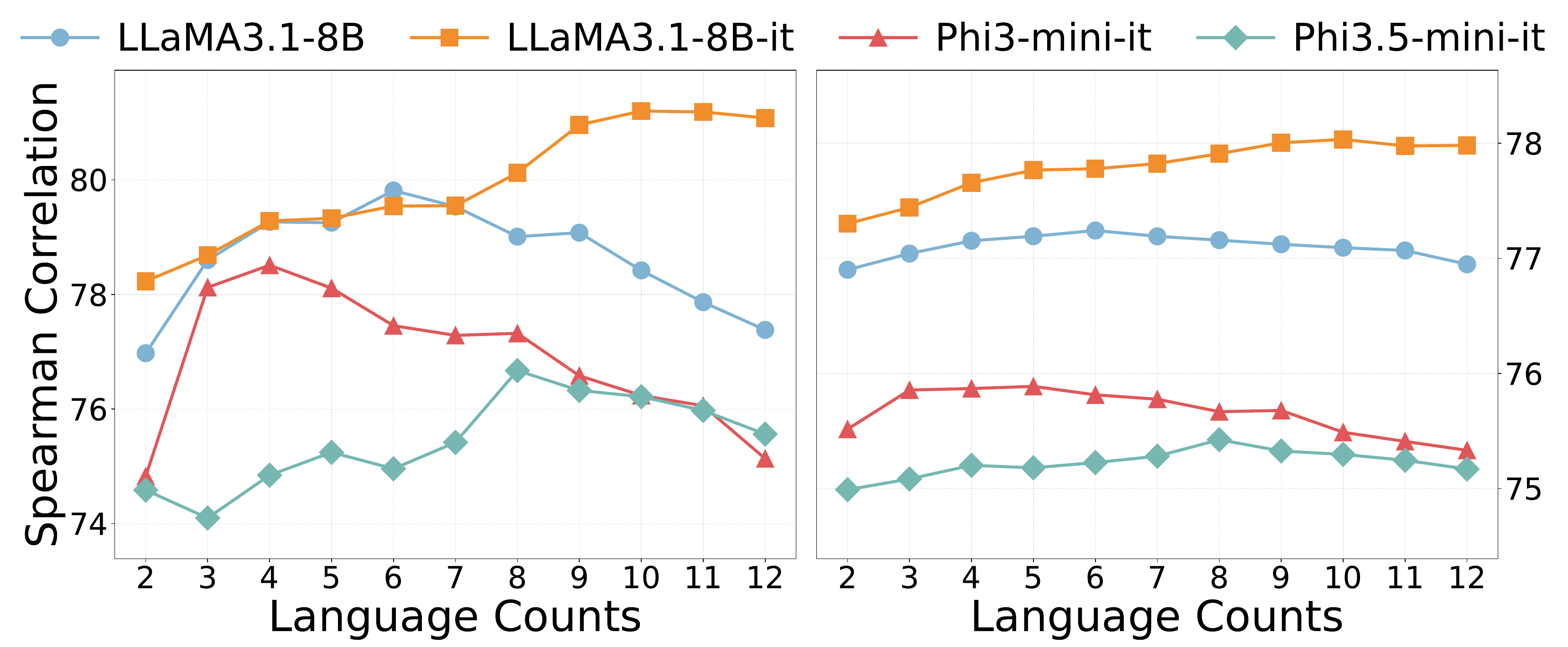}
  \caption{Analysis of language selection strategies on language-specific prompt with mean vector (left) and source vector (right) on remaining models. Comparison across different top-$m$ language selections.}
  \label{fig:analyse2_language-mean_language-source_curves_excluded_models}
\end{figure}

\begin{figure}[ht]
  \centering
  \includegraphics[width=1\columnwidth]{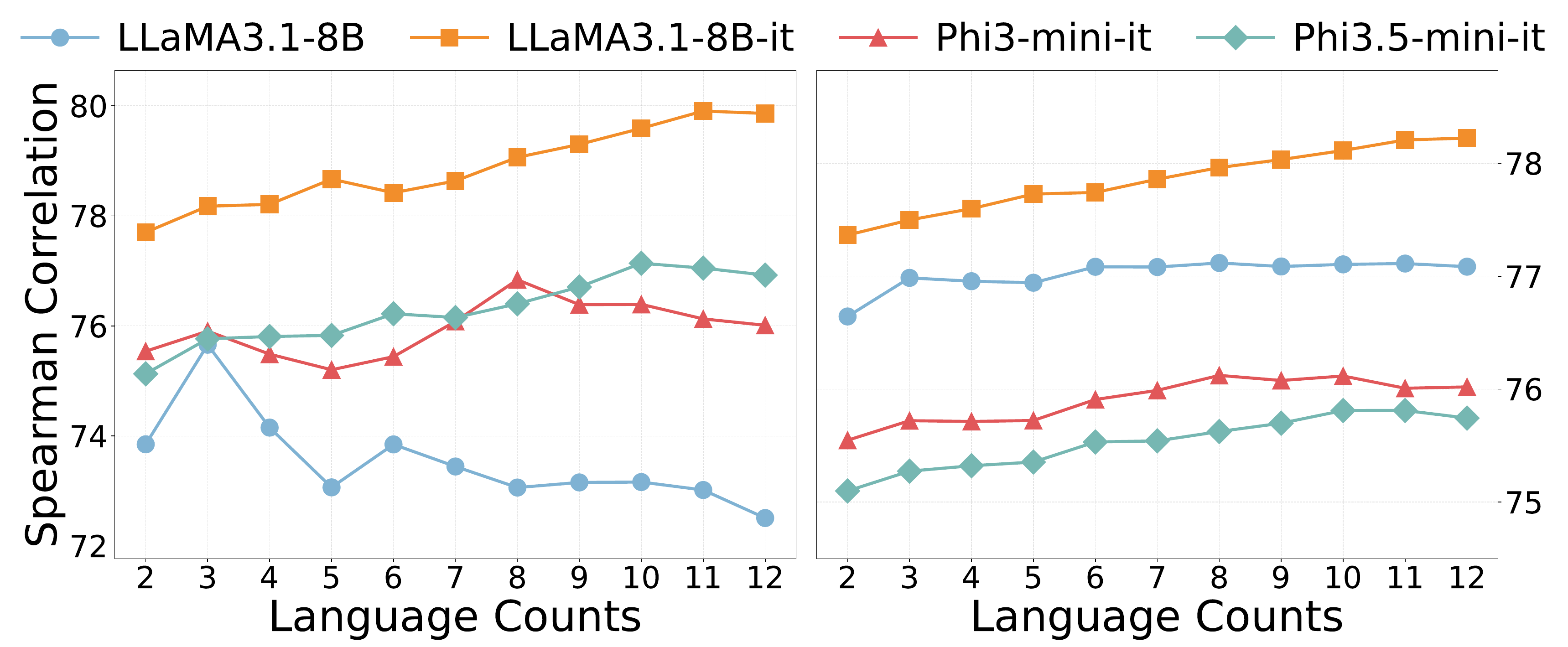}
  \caption{Analysis of language selection strategies on English prompt with mean vector (left) and source vector (right) on remaining models. Comparison across different top-$m$ language selections.}
  \label{fig:analyse2_english-mean_english-source_curves_excluded_models}
\end{figure}

Figures~\ref{fig:analyse2_language-mean_language-source_curves_excluded_models} and~\ref{fig:analyse2_english-mean_english-source_curves_excluded_models} show the same comparisons on the remaining models. The language-specific prompt with mean vector again exhibits the strongest dependence on $m$: Phi3-mini-it peaks early at $m=4$, LLaMA3.1-8B peaks around $m=6$, while LLaMA3.1-8B-it benefits from a larger pool and peaks around $m=10$. The English prompt with source vector and the language-specific prompt with source vector are comparatively stable, with only small changes as $m$ increases. These additional results reinforce the main conclusion that careful language selection is most important when using mean-vector aggregation, while source-vector settings are less sensitive but still show model-dependent optima.

\subsection{Impacts of Prompts and Semantic Vector}
\label{app:more_result_analyse3}

\begin{figure}[ht]
  \centering
  \includegraphics[width=1\columnwidth]{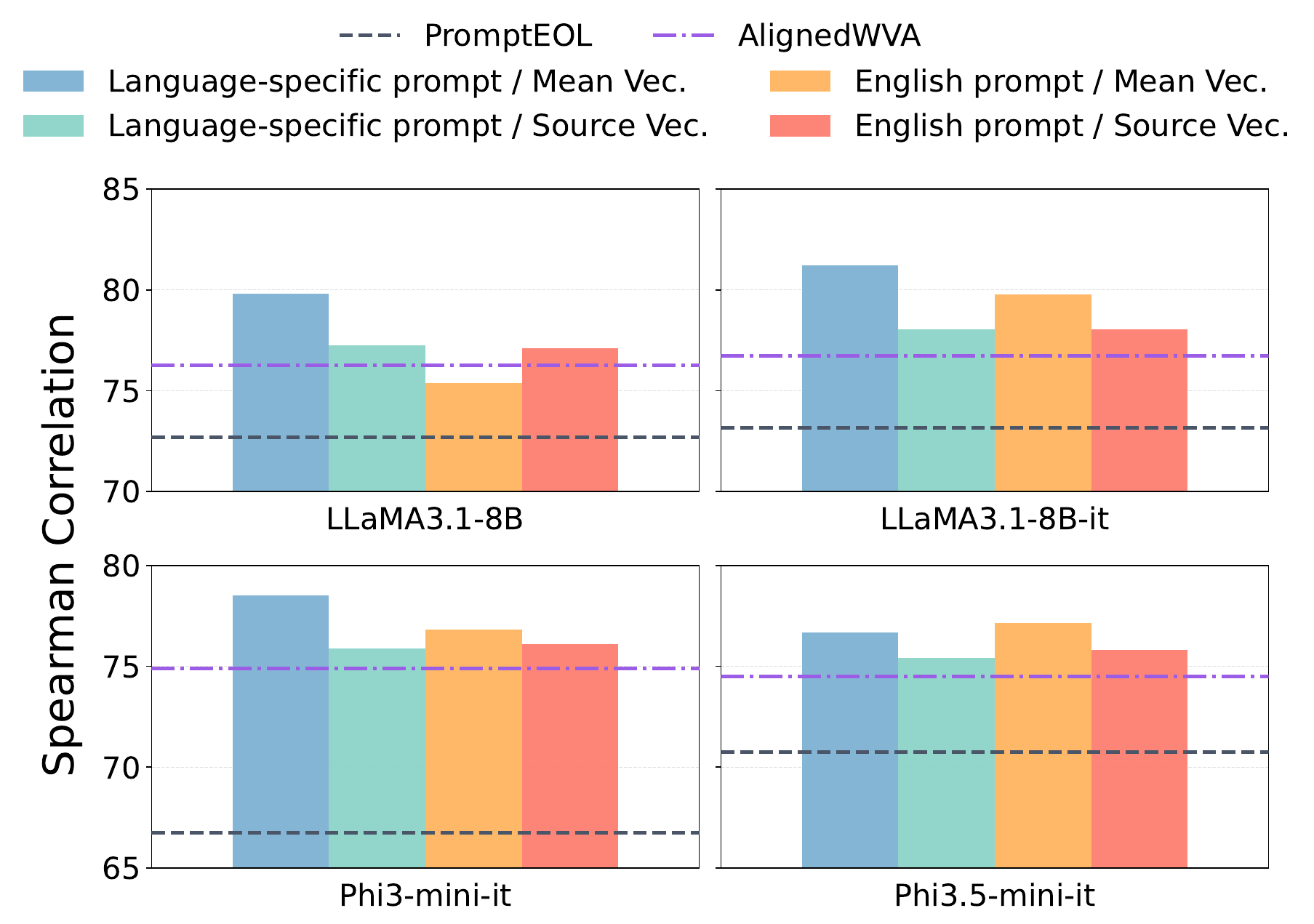}
  \caption{Comparison of English and language-specific prompts under source and mean vectors settings on remaining models.}
  \label{fig:analyse3_bars_remaining_models}
\end{figure}

\begin{figure*}[ht]
  \centering
  \includegraphics[width=1\textwidth]{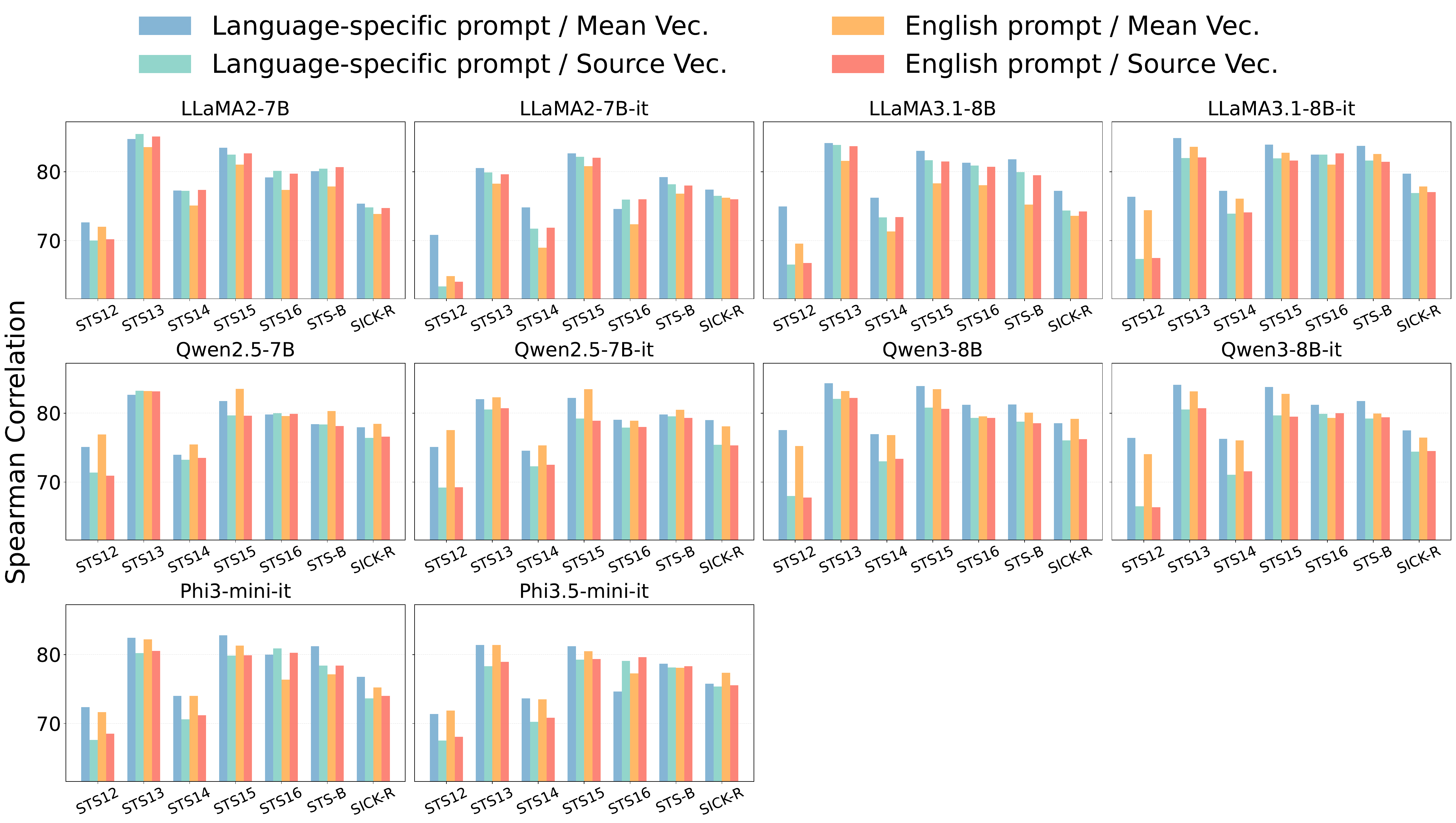}
  \caption{Comparison of English and language-specific prompts under source and mean vectors settings across STS tasks.}
  \label{fig:analyse5_task_bars}
\end{figure*}

\begin{figure*}[ht]
  \centering
  \includegraphics[width=1\textwidth]{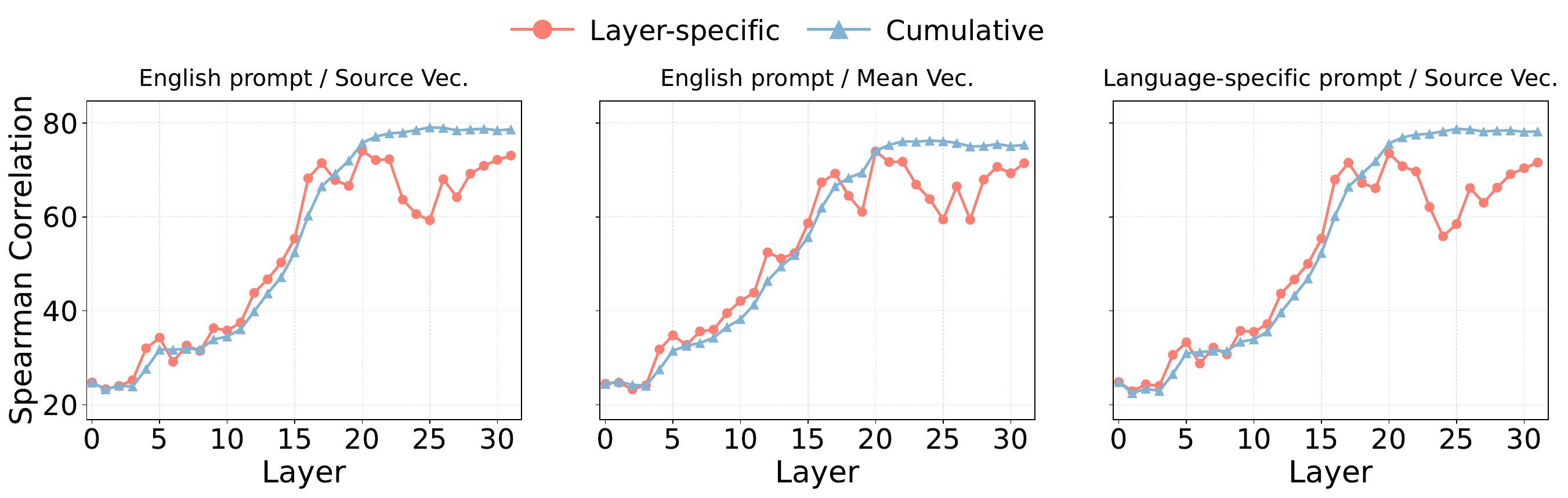}
  \caption{Average performance across tasks on LLaMA2-7B with remaining setting. Comparison between layer-specific attention and cumulative attention.}
  \label{fig:analyse4_llama2_remaining_combo_curves}
\end{figure*}

\begin{figure*}[ht]
  \centering
  \includegraphics[width=1\textwidth]{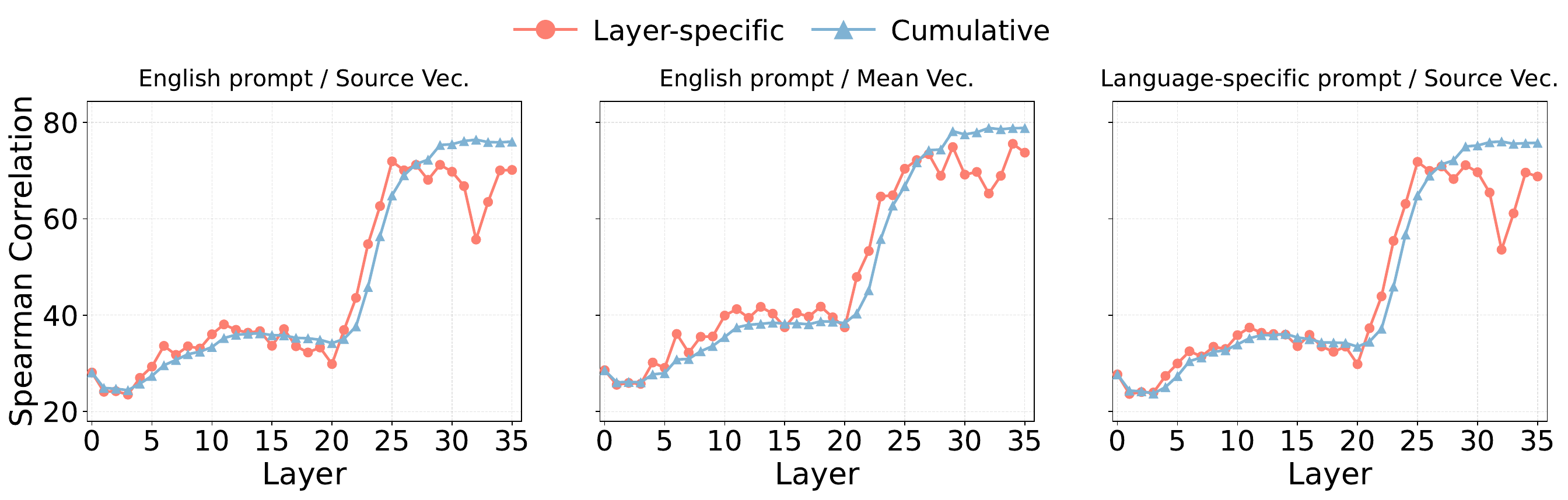}
  \caption{Average performance across tasks on Qwen3-8B-it with remaining settings. Comparison between layer-specific attention and cumulative attention.}
  \label{fig:analyse4_qwen3-8b_remaining_combo_curves}
\end{figure*}

\begin{figure*}[ht]
  \centering
  \includegraphics[width=1\textwidth]{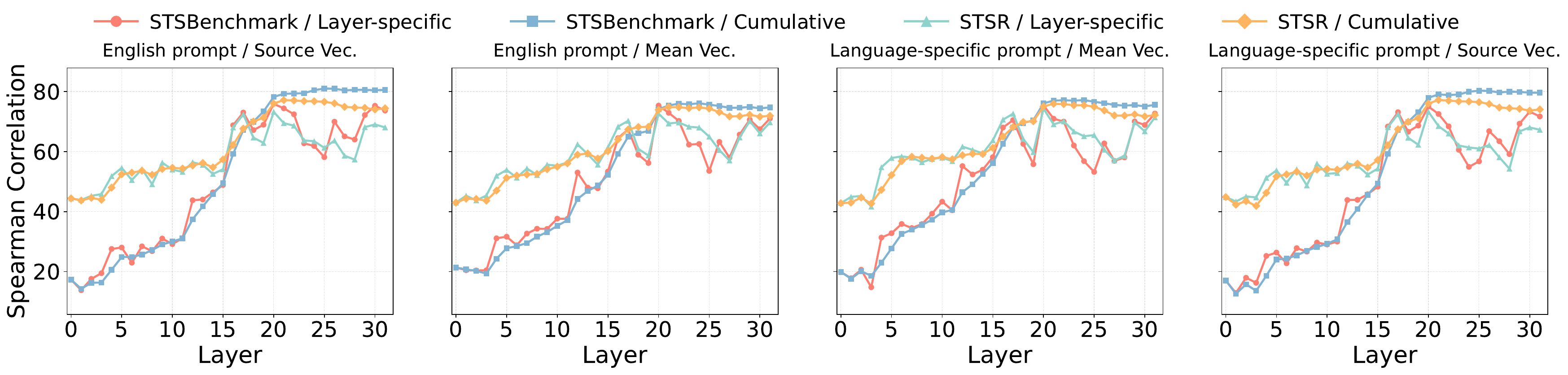}
  \caption{Task-wise performance on STSBenchmark and SICK-R for LLaMA2-7B with four settings. Comparison between layer-specific attention and cumulative attention.}
  \label{fig:analyse4_llama2_task_focus_curves}
\end{figure*}

\begin{figure*}[ht]
  \centering
  \includegraphics[width=1\textwidth]{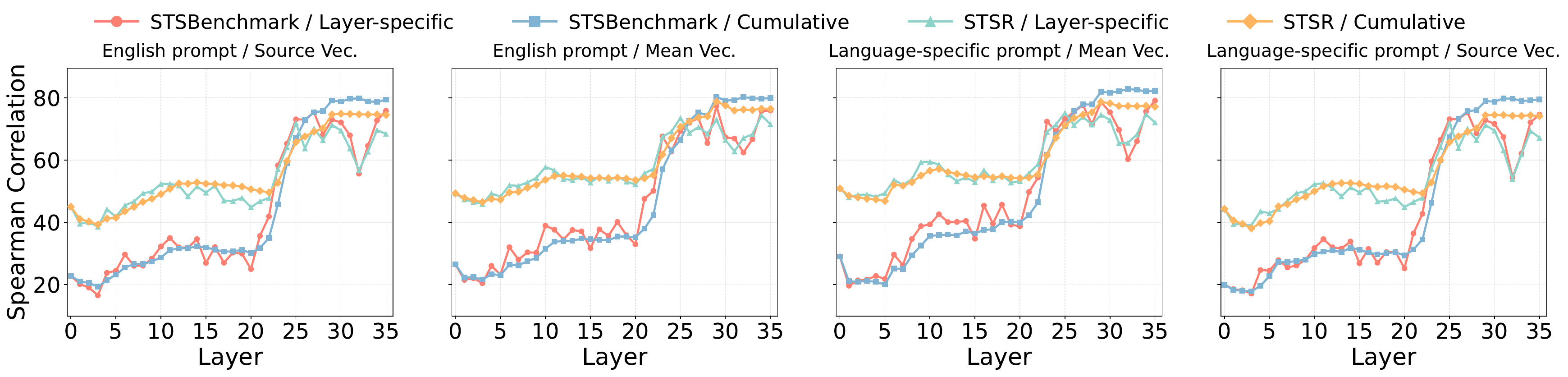}
  \caption{Task-wise performance on STSBenchmark and SICK-R for Qwen3-8B-it with four settings. Comparison between layer-specific attention and cumulative attention.}
  \label{fig:analyse4_qwen3-8b_task_focus_curves}
\end{figure*}

Figure~\ref{fig:analyse3_bars_remaining_models} reports the remaining four models. The same pattern is largely preserved: the language-specific prompt with the mean vector obtains the best result on three of the four models, including LLaMA3.1-8B, LLaMA3.1-8B-it, and Phi3-mini-it. Averaged over these remaining models, this setting reaches 79.05, outperforming the English-prompt mean-vector setting by 1.77 points and the language-specific source-vector setting by 2.40 points. The only exception is Phi3.5-mini-it, where the English prompt with the mean vector is slightly better by 0.47 points, suggesting that the advantage of localized prompts is not uniform across all model families.

Figure~\ref{fig:analyse5_task_bars} further breaks down the comparison across STS tasks. The language-specific prompt with the mean vector achieves the highest average score on six of the seven evaluated tasks, while the language-specific source vector setting is best only on STS16. Across tasks, language-specific prompts improve the mean vector by 1.24 points on average, with consistent gains on STS12, STS13, STS14, STS15, STSBenchmark, and SICK-R. In contrast, under the source vector setting, the effect of prompt language remains marginal, with task-level differences staying within 0.30 points. These results reinforce the main finding that prompt localization is most beneficial when combined with mean-vector construction, where multilingual representations are explicitly aggregated before scoring.

\subsection{Attention Aggregation Strategies}
\label{app:more_result_analyse4}

Figures~\ref{fig:analyse4_llama2_remaining_combo_curves} and~\ref{fig:analyse4_qwen3-8b_remaining_combo_curves} provide supplementary comparisons across the remaining prompt and vector configurations. The trends observed in the main text remain consistent: cumulative attention systematically achieves a higher peak average performance compared to layer-specific attention across all evaluated settings. Specifically, LLaMA2-7B shows an improvement of 4.97 points when using the English prompt combined with source vectors, 2.31 points with the English prompt and mean vectors, and 5.17 points under the language-specific prompt combined with source vectors. Correspondingly, Qwen3-8B-it exhibits gains of 4.50, 3.29, and 4.21 points under the same respective configurations. These results demonstrate that the advantages of cumulative aggregation are robust and independent of specific prompt choices or value-vector constructions.

Figures~\ref{fig:analyse4_llama2_task_focus_curves} and~\ref{fig:analyse4_qwen3-8b_task_focus_curves} further detail the performance of both strategies on the STSBenchmark and SICK-R tasks. Cumulative attention consistently yields superior task-specific peaks across nearly all settings. For instance, when applying the English prompt with source vectors to LLaMA2-7B, cumulative attention boosts performance on STSBenchmark from 75.89 to 80.99, and on SICK-R from 73.22 to 77.12. Similarly, employing the language-specific prompt with mean vectors on Qwen3-8B-it elevates STSBenchmark from 79.06 to 82.82, and SICK-R from 75.19 to 78.68. Ultimately, these supplementary findings reinforce the conclusion that cumulative attention more effectively preserves valuable cross-layer semantic signals, generating more stable representations than those derived from any single layer.

\begin{figure}[!htbp]
  \centering
  \includegraphics[width=1\columnwidth]{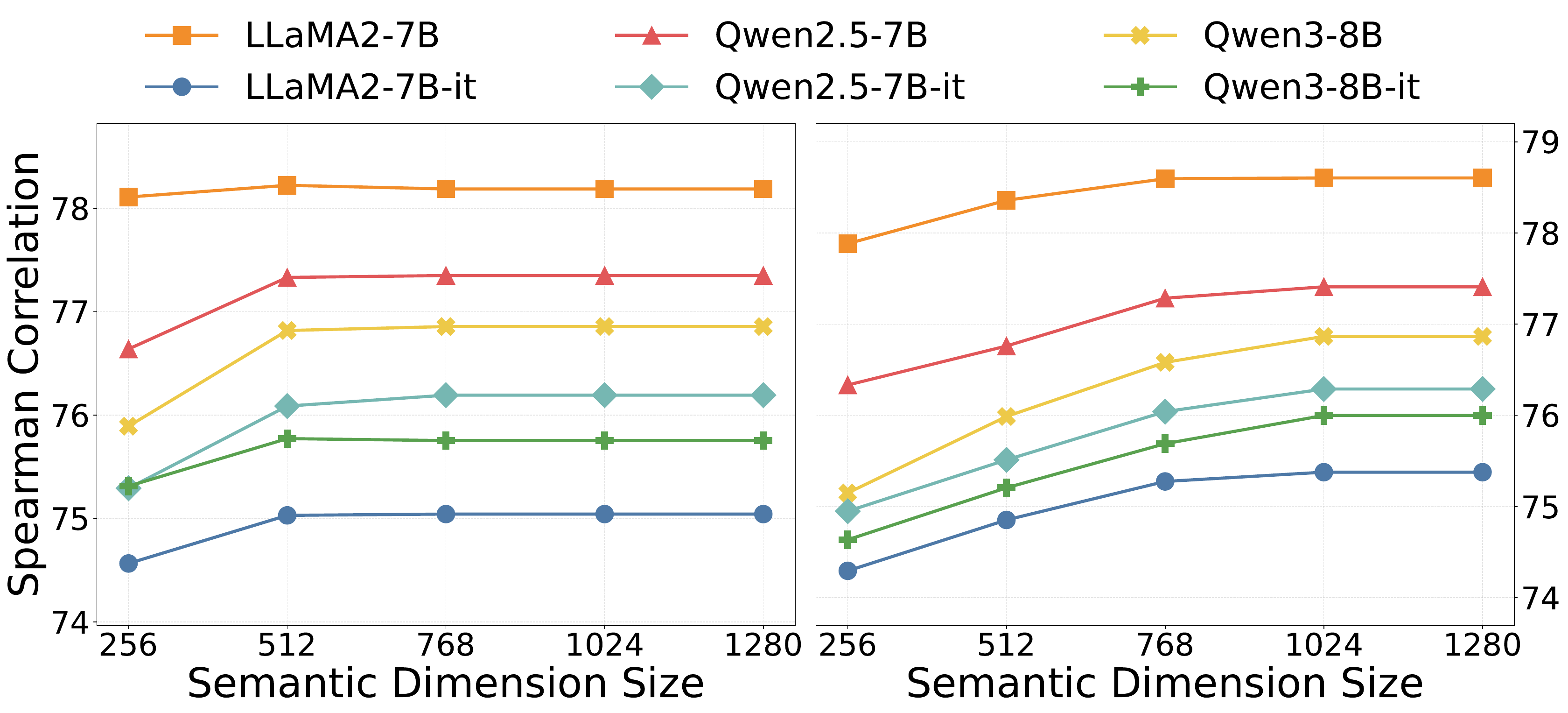}
  \caption{Analysis of semantic dimension sizes on Language-specific prompt (left) and English prompt (right) with source vector. Comparison across different top-$k$ semantic dimension selections.}
  \label{fig:analyse5_language-source_english-source_curves}
\end{figure}

\begin{figure}[!htbp]
  \centering
  \includegraphics[width=1\columnwidth]{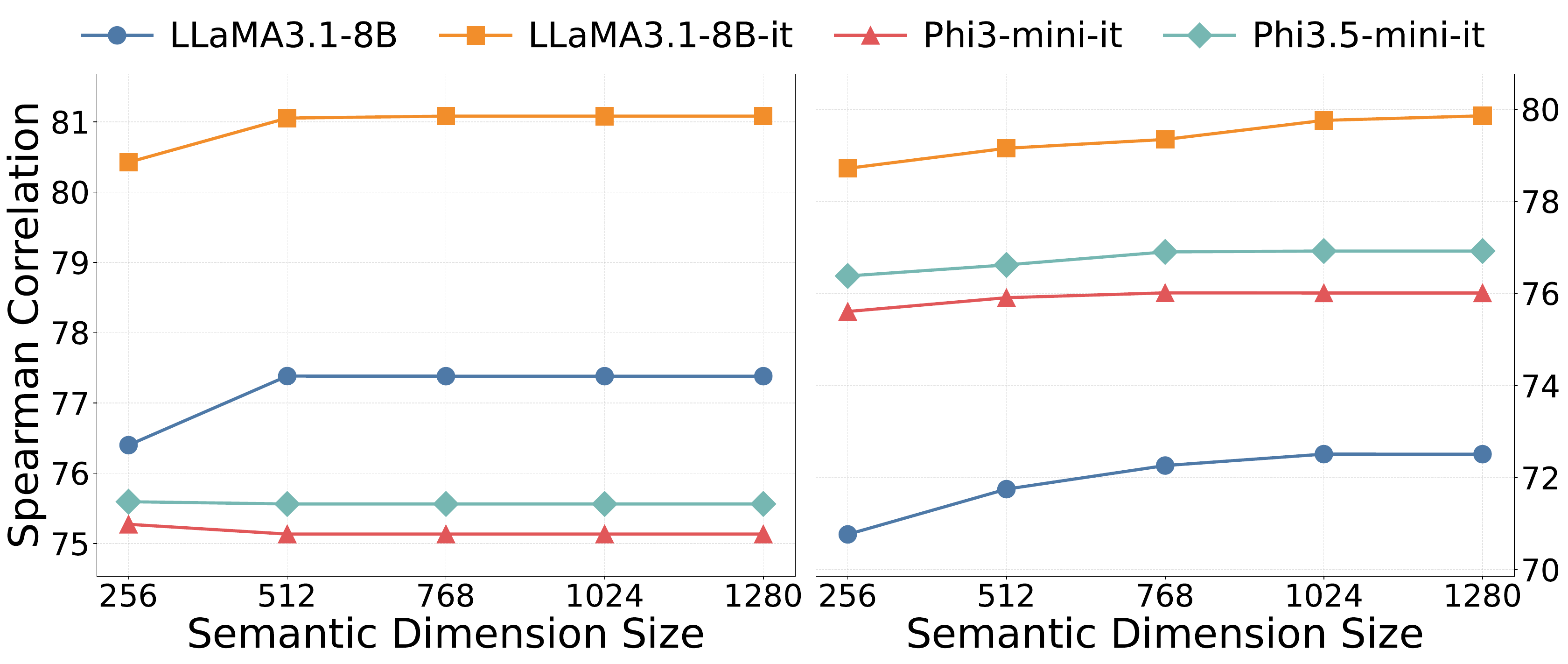}
  \caption{Analysis of semantic dimension sizes on Language-specific prompt (left) and English prompt (right) with mean vector on remaining models. Comparison across different top-$k$ semantic dimension selections.}
  \label{fig:analyse5_language-mean_english-mean_curves_excluded_models}
\end{figure}

\begin{figure}[!htbp]
  \centering
  \includegraphics[width=1\columnwidth]{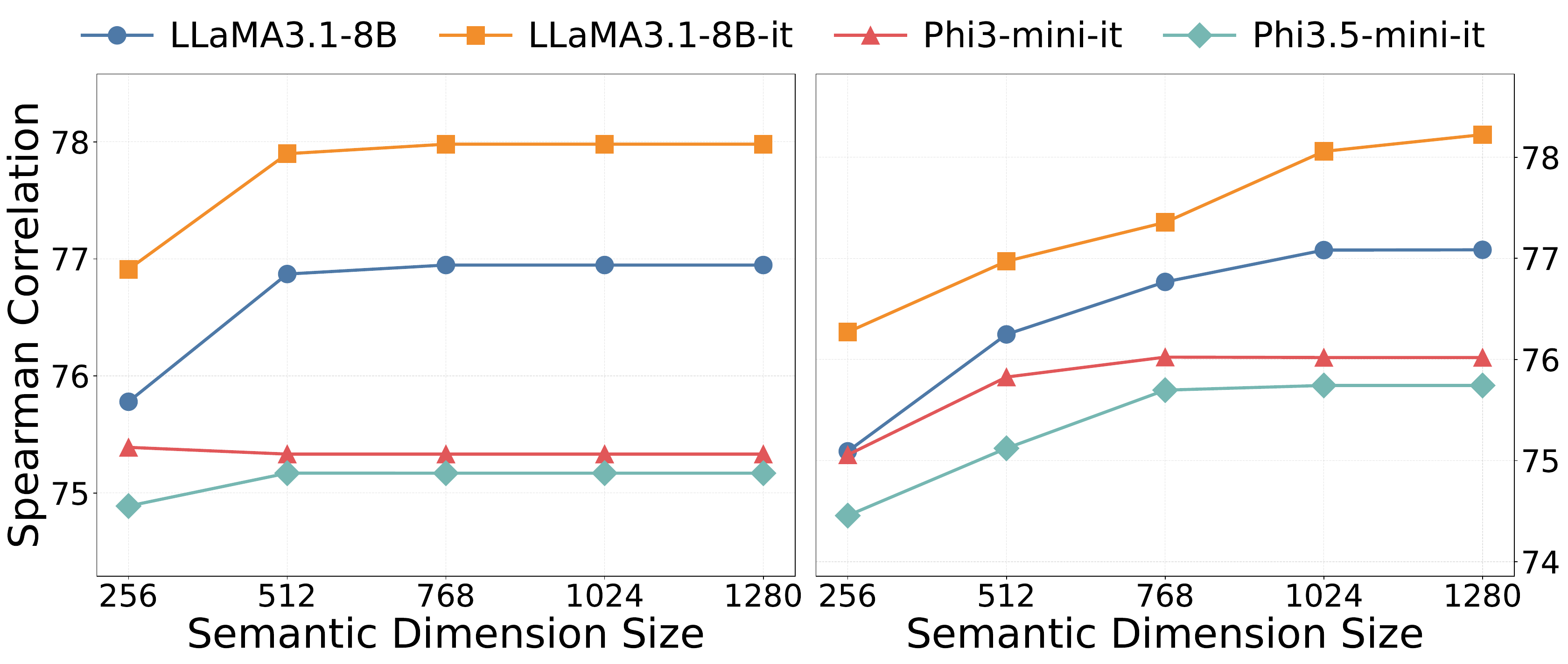}
  \caption{Analysis of semantic dimension sizes on Language-specific prompt (left) and English prompt (right) with source vector on remaining models. Comparison across different top-$k$ semantic dimension selections.}
  \label{fig:analyse5_language-source_english-source_curves_excluded_models}
\end{figure}

\subsection{Impacts Semantic Dimension Sizes}
\label{app:more_result_analyse5}

We provide additional results for the remaining prompt and vector configurations in Figure~\ref{fig:analyse5_language-source_english-source_curves}. The same overall conclusion holds: the retained dimension budget is primarily controlled by the prompt language. For the English prompt combined with a mean vector, performance increases steadily from $k=256$ to $k=1024$, with only marginal change afterwards. This indicates that English prompts still benefit from a relatively larger semantic subspace even when the vector is averaged across languages. In contrast, the configuration using a language-specific prompt combined with a source vector reaches near-optimal performance at $k=512$ or $k=768$ and then remains almost unchanged. This further supports that language-aligned prompts allow semantic information to be concentrated in a compact set of dimensions.

Figures~\ref{fig:analyse5_language-source_english-source_curves_excluded_models} and~\ref{fig:analyse5_language-source_english-source_curves_excluded_models} show the same analysis on the remaining models. Although the absolute scores vary across model families, the trend is consistent. Configurations using language-specific prompts saturate quickly, usually after $k=512$ or $k=768$, while configurations relying on English prompts show more gradual improvements as $k$ increases and often perform best at $k=1024$ or $k=1280$. These results confirm that the observed dimension-size behavior is not specific to the main set of models, but generalizes across additional model families.

\section{Effectiveness Across Different Prompts}

\begin{table*}[t!]
  \centering
  \small
  \renewcommand{\arraystretch}{1.1}
  \resizebox{\textwidth}{!}{%
    \begin{tabular}{c l c *{8}{c}}
      \toprule
      \multicolumn{2}{c}{\textbf{Methods}} & \textbf{\#Dim.} & \textbf{STS12} & \textbf{STS13} & \textbf{STS14} & \textbf{STS15} & \textbf{STS16} & \textbf{\mbox{STS-B}} & \textbf{\mbox{SICK-R}} & \textbf{Avg.} \\
      \midrule
      \multirow{8}{*}{\modelname{LLaMA2-7B}}
      & \textbf{Pretended CoT ~\cite{ck}} & 4096 & 61.00 & 79.62 & 69.42 & 74.78 & 77.20 & 72.11 & 70.56 & 72.10 \\
      & \textbf{Pretended CoT + AlignedWVA \cite{zhang2026llm}} & 4096 & 69.81 & 86.37 & 77.83 & 83.63 & \textbf{82.80} & \underline{82.07} & 76.10 & 79.80 \\
      & \ourscell{\textbf{Pretended CoT + (English Prompt, Ours)}} & \ourscell{$\phantom{^{*}}1294^{*}$} & \ourscell{\underline{71.11}} & \ourscell{\textbf{86.67}} & \ourscell{\textbf{79.05}} & \ourscell{\textbf{84.21}} & \ourscell{\underline{82.70}} & \ourscell{\textbf{82.56}} & \ourscell{\textbf{76.51}} & \ourscell{\textbf{80.40}} \\
      & \ourscell{\textbf{Pretended CoT + (Language-specific Prompt, Ours)}} & \ourscell{$\phantom{^{*}}999^{*}$} & \ourscell{\textbf{71.86}} & \ourscell{\underline{86.55}} & \ourscell{\underline{78.66}} & \ourscell{\underline{84.15}} & \ourscell{82.38} & \ourscell{81.92} & \ourscell{\underline{76.29}} & \ourscell{\underline{80.26}} \\
      \cmidrule(lr){2-11}
      & \textbf{Knowledge Enhancement ~\cite{ck}} & 4096 & 61.94 & 80.37 & 71.40 & 78.43 & 77.24 & 76.57 & 75.77 & 74.53 \\
      & \textbf{Knowledge Enhancement + AlignedWVA \cite{zhang2026llm}} & 4096 & 68.12 & 86.13 & 77.29 & 83.02 & \textbf{81.84} & 83.11 & 78.39 & 79.70 \\
      & \ourscell{\textbf{Knowledge Enhancement + (English Prompt, Ours)}} & \ourscell{$\phantom{^{*}}1117^{*}$} & \ourscell{\underline{70.69}} & \ourscell{\underline{86.45}} & \ourscell{\textbf{78.98}} & \ourscell{\textbf{84.14}} & \ourscell{81.80} & \ourscell{\textbf{83.91}} & \ourscell{\textbf{78.79}} & \ourscell{\textbf{80.68}} \\
      & \ourscell{\textbf{Knowledge Enhancement + (Language-specific Prompt, Ours)}} & \ourscell{$\phantom{^{*}}908^{*}$} & \ourscell{\textbf{70.90}} & \ourscell{\textbf{86.57}} & \ourscell{\underline{78.50}} & \ourscell{\underline{83.72}} & \ourscell{\underline{81.80}} & \ourscell{\underline{83.46}} & \ourscell{\underline{78.70}} & \ourscell{\underline{80.52}} \\
      \midrule
      \multirow{8}{*}{\modelname{LLaMA2-7B-it}}
      & \textbf{Pretended CoT ~\cite{ck}} & 4096 & 61.39 & 79.31 & 69.36 & 76.45 & 72.62 & 73.47 & 70.62 & 71.89 \\
      & \textbf{Pretended CoT + AlignedWVA \cite{zhang2026llm}} & 4096 & 62.37 & \underline{80.45} & 71.17 & 81.02 & \textbf{76.26} & 77.86 & 76.05 & 75.02 \\
      & \ourscell{\textbf{Pretended CoT + (English Prompt, Ours)}} & \ourscell{$\phantom{^{*}}1132^{*}$} & \ourscell{\underline{64.51}} & \ourscell{80.17} & \ourscell{\underline{72.85}} & \ourscell{\underline{81.98}} & \ourscell{\underline{76.11}} & \ourscell{\underline{78.74}} & \ourscell{\underline{76.59}} & \ourscell{\underline{75.85}} \\
      & \ourscell{\textbf{Pretended CoT + (Language-specific Prompt, Ours)}} & \ourscell{$\phantom{^{*}}837^{*}$} & \ourscell{\textbf{72.22}} & \ourscell{\textbf{81.73}} & \ourscell{\textbf{76.03}} & \ourscell{\textbf{83.14}} & \ourscell{75.17} & \ourscell{\textbf{79.95}} & \ourscell{\textbf{76.72}} & \ourscell{\textbf{77.85}} \\
      \cmidrule(lr){2-11}
      & \textbf{Knowledge Enhancement ~\cite{ck}} & 4096 & 57.55 & 79.32 & 68.91 & 74.36 & 74.06 & 73.18 & 72.63 & 71.43 \\
      & \textbf{Knowledge Enhancement + AlignedWVA \cite{zhang2026llm}} & 4096 & 62.09 & \underline{82.77} & 71.96 & 79.34 & \textbf{77.62} & 77.51 & 75.63 & 75.27 \\
      & \ourscell{\textbf{Knowledge Enhancement + (English Prompt, Ours)}} & \ourscell{$\phantom{^{*}}1034^{*}$} & \ourscell{\textbf{65.92}} & \ourscell{82.68} & \ourscell{\textbf{74.25}} & \ourscell{\textbf{81.27}} & \ourscell{\underline{77.49}} & \ourscell{\textbf{78.71}} & \ourscell{\underline{76.15}} & \ourscell{\textbf{76.64}} \\
      & \ourscell{\textbf{Knowledge Enhancement + (Language-specific Prompt, Ours)}} & \ourscell{$\phantom{^{*}}946^{*}$} & \ourscell{\underline{64.85}} & \ourscell{\textbf{83.15}} & \ourscell{\underline{73.51}} & \ourscell{\underline{80.49}} & \ourscell{77.39} & \ourscell{\underline{78.31}} & \ourscell{\textbf{76.22}} & \ourscell{\underline{76.27}} \\
      \midrule
      \multirow{8}{*}{\modelname{Qwen3-8B}}
      & \textbf{Pretended CoT ~\cite{ck}} & 4096 & 56.78 & 73.17 & 58.71 & 67.99 & 73.21 & 68.71 & 61.99 & 65.79 \\
      & \textbf{Pretended CoT + AlignedWVA \cite{zhang2026llm}} & 4096 & 59.12 & 79.37 & 68.50 & 73.34 & 78.85 & 73.30 & 70.53 & 71.86 \\
      & \ourscell{\textbf{Pretended CoT + (English Prompt, Ours)}} & \ourscell{$\phantom{^{*}}1007^{*}$} & \ourscell{\underline{73.02}} & \ourscell{\underline{83.26}} & \ourscell{\underline{75.77}} & \ourscell{\underline{82.05}} & \ourscell{\underline{79.02}} & \ourscell{\underline{78.57}} & \ourscell{\underline{77.97}} & \ourscell{\underline{78.52}} \\
      & \ourscell{\textbf{Pretended CoT + (Language-specific Prompt, Ours)}} & \ourscell{$\phantom{^{*}}752^{*}$} & \ourscell{\textbf{76.40}} & \ourscell{\textbf{84.42}} & \ourscell{\textbf{76.56}} & \ourscell{\textbf{83.68}} & \ourscell{\textbf{81.64}} & \ourscell{\textbf{80.63}} & \ourscell{\textbf{78.43}} & \ourscell{\textbf{80.25}} \\
      \cmidrule(lr){2-11}
      & \textbf{Knowledge Enhancement ~\cite{ck}} & 4096 & 58.28 & 79.15 & 67.73 & 75.00 & 77.14 & 72.36 & 72.37 & 71.72 \\
      & \textbf{Knowledge Enhancement + AlignedWVA \cite{zhang2026llm}} & 4096 & 64.89 & 82.16 & 73.87 & 80.11 & \textbf{79.33} & 72.61 & 73.51 & 75.21 \\
      & \ourscell{\textbf{Knowledge Enhancement + (English Prompt, Ours)}} & \ourscell{$\phantom{^{*}}914^{*}$} & \ourscell{\textbf{73.22}} & \ourscell{\underline{82.97}} & \ourscell{\underline{75.45}} & \ourscell{\textbf{81.43}} & \ourscell{78.82} & \ourscell{\textbf{76.93}} & \ourscell{\textbf{77.75}} & \ourscell{\textbf{78.08}} \\
      & \ourscell{\textbf{Knowledge Enhancement + (Language-specific Prompt, Ours)}} & \ourscell{$\phantom{^{*}}696^{*}$} & \ourscell{\underline{67.96}} & \ourscell{\textbf{83.35}} & \ourscell{\textbf{75.47}} & \ourscell{\underline{81.12}} & \ourscell{\underline{79.13}} & \ourscell{\underline{74.58}} & \ourscell{\underline{74.90}} & \ourscell{\underline{76.65}} \\
      \midrule
      \multirow{8}{*}{\modelname{Qwen3-8B-it}}
      & \textbf{Pretended CoT ~\cite{ck}} & 4096 & 52.49 & 64.14 & 54.19 & 70.53 & 74.32 & 71.10 & 62.90 & 64.24 \\
      & \textbf{Pretended CoT + AlignedWVA \cite{zhang2026llm}} & 4096 & 64.64 & 77.96 & 68.67 & 79.52 & 80.01 & 78.01 & 72.99 & 74.54 \\
      & \ourscell{\textbf{Pretended CoT + (English Prompt, Ours)}} & \ourscell{$\phantom{^{*}}1034^{*}$} & \ourscell{\textbf{75.64}} & \ourscell{\underline{83.50}} & \ourscell{\textbf{76.86}} & \ourscell{\textbf{84.16}} & \ourscell{\underline{81.03}} & \ourscell{\textbf{81.87}} & \ourscell{\underline{77.70}} & \ourscell{\underline{80.11}} \\
      & \ourscell{\textbf{Pretended CoT + (Language-specific Prompt, Ours)}} & \ourscell{$\phantom{^{*}}1240^{*}$} & \ourscell{\underline{75.43}} & \ourscell{\textbf{84.67}} & \ourscell{\underline{75.49}} & \ourscell{\underline{83.97}} & \ourscell{\textbf{81.95}} & \ourscell{\underline{81.64}} & \ourscell{\textbf{77.95}} & \ourscell{\textbf{80.16}} \\
      \cmidrule(lr){2-11}
      & \textbf{Knowledge Enhancement ~\cite{ck}} & 4096 & 59.61 & 73.26 & 63.36 & 74.70 & 75.37 & 73.33 & 71.16 & 70.11 \\
      & \textbf{Knowledge Enhancement + AlignedWVA \cite{zhang2026llm}} & 4096 & 66.31 & 80.97 & 72.61 & 79.44 & \underline{79.23} & 76.86 & 73.71 & 75.59 \\
      & \ourscell{\textbf{Knowledge Enhancement + (English Prompt, Ours)}} & \ourscell{$\phantom{^{*}}905^{*}$} & \ourscell{\underline{74.10}} & \ourscell{\underline{83.12}} & \ourscell{\textbf{76.22}} & \ourscell{\underline{82.67}} & \ourscell{\textbf{79.45}} & \ourscell{\underline{79.91}} & \ourscell{\underline{77.04}} & \ourscell{\underline{78.93}} \\
      & \ourscell{\textbf{Knowledge Enhancement + (Language-specific Prompt, Ours)}} & \ourscell{$\phantom{^{*}}683^{*}$} & \ourscell{\textbf{74.40}} & \ourscell{\textbf{83.56}} & \ourscell{\underline{76.19}} & \ourscell{\textbf{82.84}} & \ourscell{79.20} & \ourscell{\textbf{80.98}} & \ourscell{\textbf{77.25}} & \ourscell{\textbf{79.20}} \\
      \bottomrule
    \end{tabular}%
  }
  \caption{Results of Pretended CoT and Knowledge Enhancement prompts.}
  \label{tab:cot_ke_result}
\end{table*}

To further examine whether the identified semantic components are robust to different prompting strategies, we evaluate our method under two other commonly used prompts, namely \textbf{Pretended CoT} and \textbf{Knowledge Enhancement} ~\cite{ck}. The results are reported in Table~\ref{tab:cot_ke_result}. Overall, our method consistently achieves the best or second-best average performance across different models and prompts, showing that the semantic information uncovered by \textsc{DySem} is not tied to a specific prompt format.

Under the Pretended CoT prompt, \textsc{DySem} brings consistent improvements over both the original prompt-based representations and AlignedWVA. On LLaMA2-7B, our method achieves average scores of 80.40 and 80.26 with English and language-specific prompts, respectively, outperforming AlignedWVA by 0.60 and 0.46 points. The gains are more pronounced on Qwen3-8B, where the language-specific prompt version reaches an average score of 80.25, improving over AlignedWVA by 8.39 points. Similar improvements are also observed on instruction-tuned models. For example, on Qwen3-8B-it, our method obtains average scores of 80.11 and 80.16, surpassing AlignedWVA by more than 5.5 points.

Under the Knowledge Enhancement prompt, our method also maintains strong performance. On LLaMA2-7B, \textsc{DySem} achieves the best average score of 80.68 with the English prompt and 80.52 with the language-specific prompt, both outperforming AlignedWVA. On Qwen3-8B, our English-prompt variant obtains an average score of 78.08, exceeding AlignedWVA by 2.87 points. For instruction-tuned models, the improvements remain stable: on LLaMA2-7B-it, our method reaches 76.64 and 76.27 under the two prompt variants, while on Qwen3-8B-it it achieves 78.93 and 79.20, clearly outperforming the corresponding AlignedWVA results.

These results demonstrate that \textsc{DySem} is effective across different prompt designs. Although the absolute performance may vary between English and language-specific prompts, our method consistently improves STS performance over prompt-only representations and AlignedWVA. This suggests that the semantic subspaces discovered by \textsc{DySem} capture prompt-agnostic semantic knowledge inside LLMs, enabling more reliable semantic computation under diverse prompting conditions.

\section{Transferability to MTEB Datasets}
\label{app:retrieval_clustering}

To examine whether the semantic subspaces identified by \textsc{DySem} are useful beyond pairwise semantic textual similarity, we further evaluate them on representative retrieval and clustering tasks from the Massive Text Embedding Benchmark (MTEB). Specifically, we report the Spearman correlation on STS15, the average nDCG@10 score across the SciFact and ArguAna retrieval datasets, and the V-measure on the TwentyNewsgroupsClustering task. We compare \textsc{DySem} with AlignedWVA under the same models. The results are summarized in Table~\ref{tab:dysem_comparison}.

\begin{table*}[htbp]
  \centering
  \small
  \renewcommand{\arraystretch}{1.1}
  \begin{tabular}{llccc}
    \toprule
    \multirow{2}{*}{\textbf{Model}} & \multirow{2}{*}{\textbf{Method}}
    & \textbf{STS}
    & \textbf{Retrieval}
    & \textbf{Clustering} \\
    & & \textit{Spearman}
    & \textit{nDCG@10}
    & \textit{V-measure} \\
    \midrule

    \multirow{2}{*}{\textbf{LLaMA2-7B}}
    & \textbf{AlignedWVA \cite{zhang2026llm}}
    & 81.68 & 44.88 & 33.82 \\

    & \ourscell{\textbf{\textsc{DySem} (ours)}}
    & \ourscell{\textbf{83.47}}
    & \ourscell{\textbf{46.10}}
    & \ourscell{\textbf{43.83}} \\

    \midrule

    \multirow{2}{*}{\textbf{Qwen3-8B}}
    & \textbf{AlignedWVA \cite{zhang2026llm}}
    & 77.38 & 44.89 & 44.51 \\

    & \ourscell{\textbf{\textsc{DySem} (ours)}}
    & \ourscell{\textbf{83.91}}
    & \ourscell{\textbf{49.08}}
    & \ourscell{\textbf{44.63}} \\

    \bottomrule
  \end{tabular}
  \caption{Performance comparison on MTEB datasets. The best result for each model and task is shown in bold.}
  \label{tab:dysem_comparison}
\end{table*}

\textsc{DySem} consistently improves STS15 performance for both models, and the gains also transfer to retrieval and clustering tasks. On LLaMA2-7B, \textsc{DySem} improves the average retrieval score from 44.88 to 46.10, corresponding to an absolute gain of 1.22 points. It also substantially improves the TwentyNewsgroupsClustering V-measure from 33.82 to 43.83, yielding a gain of 10.01 points. These results indicate that the dynamically selected semantic dimensions capture information that is useful not only for measuring sentence-level similarity, but also for ranking relevant documents and grouping semantically related texts.

On Qwen3-8B, \textsc{DySem} achieves a 6.53-point improvement on STS15, increasing the score from 77.38 to 83.91. It also improves the average retrieval score from 44.89 to 49.08, corresponding to a gain of 4.19 points.
For clustering, \textsc{DySem} slightly outperforms AlignedWVA.
Overall, these results show that \textsc{DySem} is not limited to sentence-pair similarity. Its dynamically constructed semantic representations transfer to both retrieval and clustering, demonstrating the broader usefulness of the semantic components identified inside LLMs.

\section{Use of AI Assistants}
AI assistants were used solely for polishing purposes. They did not contribute to the generation of any core technical contents, experimental results, or research conclusions of this work.

\end{document}